\documentclass{article}

\PassOptionsToPackage{numbers, compress}{natbib}

\usepackage{amsmath}
\usepackage{amsthm}
\usepackage{stackengine}
\usepackage{graphicx}
\usepackage{caption}
\usepackage{subcaption}

\usepackage[preprint]{neurips_2023}

\usepackage[utf8]{inputenc} 
\usepackage[T1]{fontenc}    
\usepackage{hyperref}       
\usepackage{url}            
\usepackage{booktabs}       
\usepackage{amsfonts}       
\usepackage{nicefrac}       
\usepackage{microtype}      
\usepackage{xcolor}         
\usepackage{algEvalCommands}

\newtheorem{theorem}{Theorem}

\title{Streaming algorithms for evaluating noisy judges on unlabeled data - binary classification}

\author{%
 Andr\'es Corrada-Emmanuel \\
 Data Engines\\
 Real Chemistry \\
  \texttt{andres.corrada@dataengines.com}
}

\begin{document}

\maketitle

\begin{abstract}
  The evaluation of noisy binary classifiers on unlabeled data is treated as a
  streaming task - given a data sketch of the decisions by an ensemble, estimate
  the true prevalence of the labels as well as each classifier's accuracy on them.
  Two fully algebraic evaluators are constructed to do this. Both are based on the assumption that
  the classifiers make independent errors on the test items. The first is based on
  majority voting. The second, the main contribution of the paper, is guaranteed
  to be correct for independent classifiers. But how do we know the classifiers
  are error independent on any given test? This principal/agent monitoring paradox
  is ameliorated by exploiting the failures of the independent evaluator to
return sensible estimates.  Some of these failures can be traced to producing
algebraic versus real numbers while evaluating a finite test. A
  search for nearly error independent trios is empirically carried out on the 
  \texttt{adult}, \texttt{mushroom}, and \texttt{two-norm} datasets by using
these algebraic failure modes to reject potential evaluation ensembles as
too correlated. At its final steps, the searches are refined by constructing
a surface in evaluation space that must contain the true value point.
The surface comes from considering the algebra of arbitrarily correlated
classifiers and selecting a polynomial subset that is free of any correlation variables.
Candidate evaluation ensembles are then rejected if their data sketches produce
independent evaluation estimates that are too far from the constructed surface.
The results produced by the surviving evaluation ensembles can sometimes be as good as 1\%. 
But handling even small amounts of correlation remains a challenge. A Taylor expansion
of the estimates produced when error independence is assumed but the classifiers are, in fact,
slightly correlated helps clarify how the proposed independent evaluator has algebraic `blind spots'
of its own. They are points in evaluation space but the estimate of the independent evaluator
has a sensitivity inversely proportional to the distance of the true point from them.
How algebraic stream evaluation can and cannot help when done for safety or economic 
reasons is briefly discussed.
\end{abstract}

\section{Introduction}

Streaming algorithms compute sample statistics of a data stream. A \emph{data sketch}, selected
to fit the sample statistic one wants to compute, is updated every time a new item appears
in the stream. A simple example of such a streaming algorithm is the use of two
counters to compute the average value of a stream of numbers,
\begin{align}
    n, \, & \text{sum}=\sum_{i}^{n} x_i \\
\end{align}
The first integer counter tallies how many numbers have been observed so far.
The second keeps a running total of the observed values.
This data sketch is then used to compute the mean of the observed stream, 
\begin{equation}
\text{sum}/n.
\end{equation}
Two things are notable about this simple algorithm. It only uses observed
variables, and its computation is purely algebraic with them. There are no free parameters
to tune or know beforehand. This paper discusses evaluation algorithms for
binary classifiers that act similarly.

An evaluation on a finite set of labeled data is defined by the sample
statistics it computes based on knowledge of the true labels.
\emph{Unlabeled evaluation}, the problem considered here, is the
estimation of the values of these same sample statistics when the
true labels are not known. This paper looks at how to do so when
all we have are statistics of the decisions the members of an ensemble
of noisy binary classifiers make - it is a `black-box' algorithm.
There are $2^n$ possible prediction events
when we are concerned only about \emph{per-item} evaluation statistics
for $n$ binary classifiers.
Figure 1 shows how an example of a stream labeled by three classifiers.
To keep track of the decisions events at the per-item level, only 8
integer counters are needed.

\begin{figure}
  \centering
  \includegraphics[width=\textwidth]{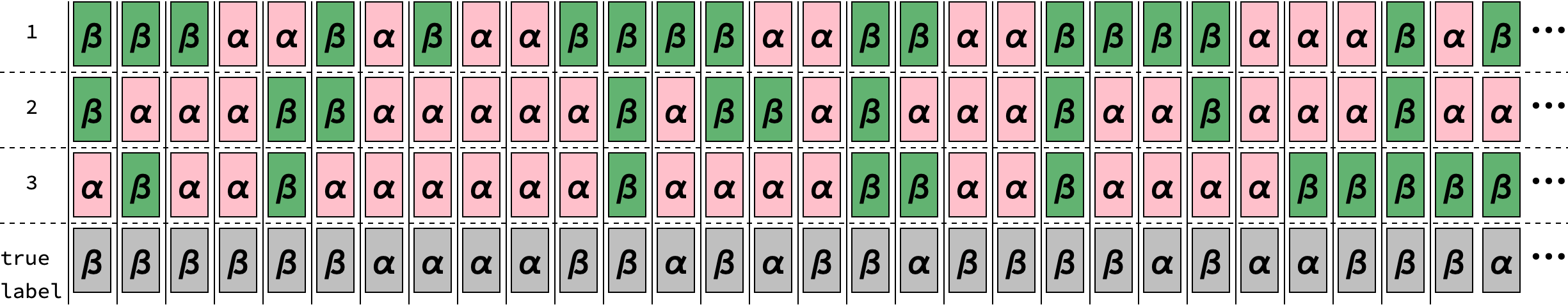}
  \caption{Stream of label predictions by three binary classifiers. Eight
  integer counters are enough to tally the number of times a particular
  prediction or voting pattern occurs when looking at \emph{per-item}
  decision events.
  }
\end{figure}

For three binary classifiers, a \emph{basic set of evaluation
statistics} is defined by the prevalence of one of the labels, say label \lbla, 
\prva, and the label accuracies, \prsa{i} and \prsb{i},
for each of the classifiers. Note that these are \emph{per item} statistics.
They cannot quantify performance across items in the stream. Theorem 1 asserts that
these variables are complete to explain the data sketches formed from
the aligned decisions of independent classifiers. If we knew the value
of the value of these evaluation statistics, we can predict exactly the
value of the per-item data sketch counters. The challenge in unlabeled
evaluation is to go the other way - to obtain estimates of the basic
evaluation statistics starting from the data sketch.

Two evaluators for binary classifiers that are fully algebraic
are built. The first is based on majority
voting (MV). It decides what the correct answer key must be for the test.
This makes it nearly impossible for it to return a correct answer even when its
assumptions are satisfied. Even worse, it always provides seemingly
correct estimates even when its assumptions are violated.
The second evaluator is fully inferential. It never decides what the true label
is for any items. Theorem 2 proves that this approach will correctly trap the true
evaluation point to just two point candidates in evaluation space. Unlike the
MV evaluator, it can return obviously incorrect estimates.

This paper assumes that these streaming evaluators are being deployed
in an environment that has a principal/agent monitoring paradox.
Evaluation ensembles, like decision ensembles, work best when they are
independent in their error. Just as one would not want to incur the
technical debt of having an ensemble of classifiers that always agreed,
it makes little sense to deploy evaluation ensembles that are highly
correlated. This raises two challenges when working on unlabeled data -
how do we find these error independent evaluation trios, and how do
we know, on any given evaluation, that they are still independent, 
or nearly so?

The approach taken here to ameliorate this monitoring paradox is that
the failures of the independent evaluator can be used to exclude
evaluation trios that are too correlated. The independent evaluator,
by construction, is deterministic and always returns algebraic
numbers. But its answers do not always make sense. We can detect
this because we have prior knowledge about what \emph{seemingly correct}
evaluation estimates look like.

\subsection{Properties of the true evaluation point}

All the basic evaluation statistics are integer ratios by construction.
For example, \prva, must be the ratio of two integers. Its numerator
is some integer between 0 and the size of the test. The denominator,
the size of the test. By construction, their ratio lies inside
the unit interval. Similar considerations apply to any of the label
accuracies for the classifiers - their true value must be an unknown
integer ratio in the unit interval.

\emph{Seemingly correct estimates} are estimated values that seem
to be correct because they have this real, integer ratio form. 
Estimates that do not have this form are obviously incorrect.
The naive evaluator built using majority voting by the ensemble 
always returns seemingly correct answers, never alerting its users 
that it is wrong and its evaluation assumptions do not apply on a given test.

The independent evaluator constructed from Theorems 1 and 2 is not
like that. It fails, with varying degrees, to return seemingly
correct estimates when the assumptions of a test are violated.
The failures vary in their severity. The empirical hypothesis explored
here is that their severity are indicative of the magnitude of 
the unknown decision correlations that are needed to correctly
predict the observed data sketch.

\subsection{A sample definition of decision error independence}

Theorem 2 provides a closed, algebraic solution to the evaluation
variety defined by the data sketch of error independent classifiers.
The variety, the geometrical object in variable space that contains
all points that satisfy a polynomial ideal, consists of just two
points for these classifiers. But as noted above, independent
evaluation ensembles are rare.

Handling correlation correctly requires that we introduce
new evaluation variables to quantify it.  Theorem 3 provides a 
a constructive proof of how to connect the data sketch of
correlated classifiers to polynomials using these new correlation
statistics plus those in the basic evaluation set.
The evaluation variety, the set of values for the evaluation
statistics that solves the polynomials, is not solved here for
correlated classifiers. 
Nonetheless, it can be proven that the variety exists, its exact shape 
in evaluation space to be determined in future work.
Nonetheless, a partial characterization of its shape is possible
because of Theorem 3. The same process that solved the polynomial
system for independent classifiers in Theorem 2 achieves a partial
disentanglement of the variables when they are correlated. This
defines a subset of the generating set that defines a surface
computable without any knowledge of the correlations between the
classifiers. This surface is not the evaluation variety but
is guaranteed to contain it. This surface is used in the experiments
to construct nearly independent evaluation ensembles.

\subsection{Previous work and related topics}
\label{sec:previous-work}
A mathematical treatment of the correctness of the decisions made
when correct labels are assigned to majority voting dates back
to Condorcet's analysis of the correctness of human juries during
the French Revolution. But it was not until almost two centuries
later that a mathematical treatment of using juries, this time human doctors,
to evaluate themselves using only their aligned decisions was published
by Dawid and Skene \cite{Dawid79}. It proposed a probabilistic solution 
to evaluation by minimizing a likelihood function using the
EM algorithm.
This work was followed up in the early 2010s by a succession
of papers in the NeurIPS conferences that took a Bayesian approach to
constructing stream evaluator (\cite{Raykar2010}, \cite{Wauthier2011},
\cite{Liu2012}, \cite{Zhou2012}, \cite{Zhang2014}). Applications to
evaluating workers in crowd-sourcing platforms was a big motivator
for some of this research. The algebraic methodology proposed here
would be economically impractical in such applications 
since it requires all the classifiers labeling every item in the stream.
\cite{Platanios2014} and \cite{Platanios2016} have an algebraic
treatment of the stream error rate for independent functions. As discussed
in the supplement, although the general Platanios agreement equations
are correct, the solution for independent functions is incorrect as it
rests on an impossible factorization of terms. This and other mathematical
mistakes made by \cite{Platanios2014} are discussed within the algebraic
geometry framework of this paper.

The algebraic approach proposed here is closest to another probabilistic
method, one proposed by Parisi et al.\ \cite{Parisi1253}. 
Rather than minimizing a likelihood
function, it considers the spectral properties of matrices created by
moments of the observed decisions. By hypothesizing hidden distributions,
it then tries to carry out a matrix decomposition that eventually gives
evaluation estimates. In contrast, the independent evaluator proposed here 
is purely algebraic. It invokes no assumptions about distributions.
Nonetheless, Theorems 1 and 2 discussed here should be contrasted
with Theorem 1 in the paper by Jaffe et al.\ \cite{Jaffe2015}, a
solution for distribution-independent classifiers.
The Supplement details the mathematical similarities and differences
between the two solutions.

But research on direct evaluation seems to have waned since the 2010s. 
The 1st NeurIPS workshop on AI safety occurred last year and its RFP had 
a section on \emph{monitoring} and \emph{anomaly detection} that does not cite 
any of the above work \cite{Hendrycks2021}. 
Instead, research has focused on other aspects of monitoring that are important for
AI safety - for example, carrying out risk minimization computations given
unknown operating points for an ensemble as was done by Steinhardt et al.\ in 
\cite{Steinhardt2016}. 

Although stream algorithms do not usually calculate hidden knowledge statistics
in a stream, some do. Good-Turing frequency smoothing \cite{GT} is one
such algorithm. Its core intellectual idea - that you can estimate
the probability of seeing hitherto unseen types of items using only the
count for observed types - is used by LLMs whenever they want to estimate
token sequences that were never observed during training.

Finally, most of the mathematical tools used in this paper come from
\emph{algebraic geometry} (AG) \cite{Cox}. The use of algebraic concerns to
study statistical problems was pioneered by Pistone et al.\ \cite{Pistone}. 
The topic is
known as \emph{algebraic statistics}. Like most of statistics, it is
focused mostly on topics related to inferring distributions. This paper
uses AG to estimate sample statistics.

\section{Using majority voting to evaluate noisy classifiers}

The correctness of decisions made by an ensemble that majority votes (MV) depends, roughly
speaking, on two things. They must be error independent, and their
labeling accuracies must be greater than 50\% on each label. 
Item labels decided by majority voting will be correct more often
than not if these conditions are met. This section details how a
\emph{naive} algebraic stream evaluator can be built on the basis
of this decision algorithm. It works by imputing the correct labels
or \emph{answer key} for the observed items.

The integer counters of the per-item data sketch can
be trivially turned into frequency variables. Each of the counters in
a decision sketch tallies how often a decision event has
been seen in the stream so far. There are only 8 decision events when 
considering the per-item decisions of three classifiers. The sum of their
fractional frequencies, \freqthree{\ell_1}{\ell_2}{\ell_3}, must sum
to one,
\begin{equation}
    \freqthree{\lbla}{\lbla}{\lbla} + \freqthree{\lbla}{\lbla}{\lblb} +
    \freqthree{\lbla}{\lblb}{\lbla} + \freqthree{\lblb}{\lbla}{\lbla} +
    \freqthree{\lblb}{\lblb}{\lbla} + \freqthree{\lblb}{\lbla}{\lblb} +
    \freqthree{\lbla}{\lblb}{\lblb} + \freqthree{\lblb}{\lblb}{\lblb} = 1.
\end{equation}

The logic of MV evaluation is straightforward. The true label is given
by MV, therefore the prevalence of a label is equal to the frequency
that label was the majority vote. The estimate for the \lbla \, label
is thus a simple linear equation of these ensemble decision frequencies,
\begin{equation}
    \prva^{(\text{MV})} = \freqthree{\lbla}{\lbla}{\lbla} + 
    \freqthree{\lbla}{\lbla}{\lblb} +
    \freqthree{\lbla}{\lblb}{\lbla} +
    \freqthree{\lblb}{\lbla}{\lbla}.
\end{equation}
Similarly, we can write down algebraic formulas of the decision frequencies
for each classifiers label accuracy. For classifier 1, those estimates are,
\begin{align}
    \prva^{(\text{MV})} &= 1 - \frac{\freqthree{\lblb}{\lbla}{\lbla}}{\mvprva} \\
    \prvb^{(\text{MV})} &= 1 - \frac{\freqthree{\lbla}{\lblb}{\lblb}}{\mvprvb}.
\end{align}
The MV evaluator considers a classifier wrong if it votes against the majority.

These are algebraic functions of the frequencies derived from the
data sketch, there are no free parameters. In addition, the estimates returned by 
them are always seemingly correct. All the MV estimates of prevalence and label
accuracies are integer ratios inside the unit interval. The MV evaluator
will never be able to alert its user that its assumptions are incorrect even
when they are wildly off the mark.

\subsection{The drawbacks of evaluating by deciding}

Decision and inference are traditionally recognized as separate concerns
in Machine Learning. Avoiding decisions and its hard choices until they 
are absolutely required typically leads to better performance. So it is
here. Making a hard choice on the true label is going to be incorrect
on some unknown fraction of the events that produced a particular voting
pattern. Some of the times the ensemble voted (\lbla, \lblb, \lbla) it
could have been a \lblb item, not an \lbla one. This is expressed by the
following equation,
\begin{equation}
    n_{\ell_1, \ell_2, \ell_3} = \#(\ell_1, \ell_2, \ell_3 \svbar \lbla) +
    \#(\ell_1, \ell_2, \ell_3 \svbar \lblb).
    \label{eq:label-decomposition}
\end{equation}
The number of times we saw a voting pattern is equal to the sum of times
the items were \lbla \, plus the times it was \lblb.
For any voting pattern by the ensemble, both labels are possible for any one
item, no matter what the majority says. Zeroing out one term in this sum is
an approximation. The supplement works out how this decision step
means that the MV evaluator is hardly ever right even though it always seems
so. Fixing this naive MV evaluator is easy - include both terms when expressing
data sketch frequencies. Carrying out evaluation with these full equations
is much harder but leads to an evaluator that is guaranteed to be correct
when its assumptions are true.

\section{Fully inferential evaluation of sample independent binary classifiers}

Systems of equations can be wrong. Care must also be taken that they
they define objects that exist so as to avoid making statements about non-existing
entities. The two mathematical objects of concern here are systems of polynomial
equations and the geometrical objects consisting of the points that solve them.
The following theorem does this for error independent classifiers. It establishes
that the basic evaluation statistics are sufficient to explain all observed data
sketches created by them.

\begin{theorem} \label{th:ind-classifiers1}
The \emph{per-item} data sketch produced by independent classifiers is
complete when expressed as polynomials of variables in the basic evaluation
set. 
\begin{align}
  \freqthree{\alpha}{\alpha}{\alpha}  ={}&  \prva \prs{1}{\alpha} \prs{2}{\alpha} \prs{3}{\alpha} \, +(1 - \prva) (1 - \prs{1}{\beta}) (1 - \prs{2}{\beta}) (1 - \prs{3}{\beta})\\
  \freqthree{\alpha}{\alpha}{\beta}  ={}&  \prva  \prs{1}{\alpha} \prs{2}{\alpha} (1 - \prs{3}{\alpha}) \, +(1 - \prva) (1 - \prs{1}{\beta}) (1 - \prs{2}{\beta}) \prs{3}{\beta}\\
  \freqthree{\alpha}{\beta}{\alpha}  ={}&  \prva  \prs{1}{\alpha}  (1 - \prs{2}{\alpha})  \prs{3}{\alpha} \, + (1 - \prva) (1 - \prs{1}{\beta}) \prs{2}{\beta} (1 - \prs{3}{\beta})\\
  \freqthree{\beta}{\alpha}{\alpha}  ={}&  \prva  (1 - \prs{1}{\alpha}) \prs{2}{\alpha} \prs{3}{\alpha} \, + (1 - \prva) \prs{1}{\beta} (1 - \prs{2}{\beta}) (1 - \prs{3}{\beta})\\
  \freqthree{\beta}{\beta}{\alpha}  ={}&  \prva  (1 - \prs{1}{\alpha}) (1 - \prs{2}{\alpha}) \prs{3}{\alpha} \, + (1 - \prva) \prs{1}{\beta} \prs{2}{\beta} (1 - \prs{3}{\beta})\\
  \freqthree{\beta}{\alpha}{\beta}  ={}&  \prva  (1 - \prs{1}{\alpha})  \, \prs{2}{\alpha} \, (1 - \prs{3}{\alpha}) \, +(1 - \prva) \prs{1}{\beta} (1 - \prs{2}{\beta}) \prs{3}{\beta}\\
  \freqthree{\alpha}{\beta}{\beta}  ={}& \prva  \prs{1}{\alpha}  (1 - \prs{2}{\alpha})  (1 - \prs{3}{\alpha}) \, + (1 - \prva) (1 - \prs{1}{\beta}) \prs{2}{\beta} \prs{3}{\beta}\\
  \freqthree{\beta}{\beta}{\beta}  ={}& \prva  (1 - \prs{1}{\alpha})  (1 - \prs{2}{\alpha}) (1 - \prs{3}{\alpha}) \, + (1 - \prva) \prs{1}{\beta} \prs{2}{\beta} \prs{3}{\beta}
\end{align}
These polynomial expressions of the data sketch form a generating  set 
for a non-empty polynomial ideal, the \emph{evaluation ideal}. The \emph{evaluation variety}, 
the set of points  that satisfy all the equations in the ideal is also non-empty and contains
the true evaluation point.
\end{theorem}
\begin{proof}[Sketch of the proof]
   The assumption that true labels exist for the stream items underlies the algebraic
   work required for the proof. The correct label of each item can be encoded in
   indicator functions, \indgt{s}{\ell}, that are 1 if its argument is the correct 
   label for item s, and zero otherwise. The existence of a true label for an item
   $s$ is then equivalent to the equation,
   \begin{equation}
       \indgt{s}{\lbla} + \indgt{s}{\lblb} = 1.
   \end{equation}
   Consider now the first term in Equation~\ref{eq:label-decomposition}, \numaevents,
   as it relates to, say, the decisions event (\lbla, \lblb, \lbla). By using the 
   predicted labels by the classifiers for a given items $s$, the following
   expression is exactly equal to one precisely at those decisions events but
   zero otherwise,
   \begin{equation}
       \label{eq:ind-event}
       \indgt{s}{\ell_{1,s}} \, (1 -  \indgt{s}{\ell_{2,s}}) \, \indgt{s}{\ell_{3,s}}.
   \end{equation}
   The proof of having a complete representation using the basic evaluation variables
   then hinges in equating the average of this equation to the variables as follows,
   \begin{equation}
       \label{eq:ind-pol-event}
       \frac{1}{n_\lbla} \sum_{\indgt{s}{\lbla}=1} \indgt{s}{\ell_{1,s}} \, (1 -  \indgt{s}{\ell_{2,s}}) \, \indgt{s}{\ell_{3,s}} = \prva \, \prsa{1} \, (1-\prsa{2}) \, \prsa{3}.
   \end{equation}
   This equality is only true for independent classifiers because we have substituted the
   average of products of the indicator functions by products of their averages. 
   New correlation variables are introduced and then set to zero to define rigorously
   a sample definition of decision correlations. For example, the definition of the pair 
   correlation variable on a label is given by,
   \begin{equation}
       \label{eq:pair-correlation}
       \corrtwo{i}{j}{\ell} = \frac{1}{n_\ell} \sum_{\indgt{s}{\ell}=1} 
       (\indgt{s}{\ell_{i,s}} - \prs{i}{\ell}) \, 
        (\indgt{s}{\ell_{j,s}} - \prs{j}{\ell}) =
        \left( \frac{1}{n_\ell} \sum_{\indgt{s}{\ell}=1} \indgt{s}{\ell_{i,s}} \, \indgt{s}{\ell_{j,s}} \right)
        - \prs{i}{\ell}  \, \prs{j}{\ell}.
   \end{equation}
   Setting these pair correlations to zero then guarantees that we
   can write averages of the product of the indicator functions for
   two classifiers as the product of their label accuracies. Similar considerations
   apply to the product of the indicators for three classifiers. The consequence
   is that any decision event frequency by independent classifiers is complete when 
   written in terms of the basic evaluation statistics.
   All data sketches produced by independent classifiers
   are predicted by the basic statistics. Since the proof is constructive and
   starts from expressions for the true evaluation point, we know that there is at least one point that satisfies all these polynomial equations. We conclude that the evaluation variety for independent classifiers exists and it contains the true evaluation point.
\end{proof}

Theorem 2 details exactly what the evaluation variety for independent classifiers must be.
\begin{theorem}
The polynomial generating set for independent classifiers has an evaluation
variety that has two points, one of which is the true evaluation point.
\end{theorem}
\begin{proof}[Sketch of the proof]
The quartic polynomials of the independent generating set are not trivial 
to handle. A strategy for solving
them is to obtain algebraic consequences of them that isolate the variables.
Using the tools of AG, this can be accomplished for independent classifiers.
Solving their polynomial system is accomplished by calculating another
representation of the evaluation ideal, called the Gr\"{o}bner basis,
that does this. It can be arranged to isolate the \prva in a quadratic
\begin{equation}
    a(\ldots) \prva^2 + b(\ldots) \prva + c(\ldots) = 0.
\end{equation}
The coefficients a, b, and c are polynomials of the decision frequencies. Since
this is a quadratic, it follows from the quadratic formula that it can only
contain two solutions. This, coupled with the fact that other equations in
the evaluation ideal are linear equations relating \prva to \prsa{i} or \prsb{i}
variables leads one to conclude that only two points exist in the evaluation variety
of independent classifiers.
\end{proof}

Table~\ref{tbl:prevalence-comparison} compares the prevalence estimates of the fully
inferential independent evaluator with the naive MV one. By construction, it will
be exact when its assumptions apply. But unlike the naive MV evaluator, this formula
can return obviously wrong estimates. The next section details how one can carry out
experiments on unlabeled data to find evaluation ensembles that are going to be
nearly independent.

\begin{table}
  \caption{Algebraic evaluation formulas for the prevalence of \lbla, \prva, for three classifiers 
  making independent errors on a test. 
  The $\Delta_{i,j}$ and \freqtwo{i}{\beta} variables are
  polynomial functions of the data sketch frequency counters. Each \freqtwo{i}{\beta} is the
  frequency classifier `i' voted for the $\beta$ label. The deltas are equal to
  $\freqthree{i}{j}{\beta} - \freqtwo{i}{\beta} \, \freqtwo{j}{\beta}$, where
  \freqthree{i}{j}{\beta} is the frequency classifiers `i' and `j' voted simultaneously
  for the $\beta$ label.}
  \label{tbl:prevalence-comparison}
  \centering
  \begin{tabular}{lc}
    \toprule
    Evaluator     & \prva  \\
    \toprule
    Majority Voting & \mvprev  \\
    \midrule
    Fully inferential  &  \aeprev  \\
    \bottomrule
  \end{tabular}
\end{table}

\section{Experiments with the failure modes of the independent stream evaluator}

If perfect evaluations are not possible, one should prefer methods that alert us
when they fail or their assumptions are incorrect. The experiments discussed here
show how we can lever the self-alarming failures of the independent evaluator
to reject highly correlated evaluation ensembles. There are four failure
modes for the independent evaluator,
    \begin{itemize}
        \item The evaluation variety corresponding to the independent
        evaluation ideal is the empty set - no points in evaluation
        space can zero out the equations in the evaluation ideal.
        \item The two evaluation points contain complex numbers.
        \item The two evaluation points lie outside the real, unit
        cube.
        \item The estimated values contain unresolved square roots.
    \end{itemize}
The fourth failure mode is interesting theoretically but not as practical.
The Supplement details how an unresolved square root in the evaluation
estimates can be used to prove that the classifiers were not error independent
in the evaluation. Its theoretical interest lies in demonstrating that
algebraic numbers, unlike real ones, can be used to self-alarm, in an almost
perfect fashion, when its assumptions are violated.

The first three failure modes are more practical. The first set of experiments
will look at how the failures can be used to estimate what test sizes are
less likely to have failed evaluations when we use the independent evaluator. 
The second set of experiments profiles how
well another rejection criteria, this time based on the polynomial
generating set for correlated classifiers, can help identify nearly independent
evaluations.

\subsection{Rejecting highly correlated evaluation ensembles}

Given a set of unlabeled data and a smaller set of labeled training data,
the goal of the first set of experiments is to construct and identify nearly independent evaluation ensembles on a larger, unlabeled portion of data. 
This simple experimental set-up is meant to mimic a possible Auto-ML application 
of stream evaluation. The experiments are meant to answer the question -
how big should an evaluation test be? This is done by profiling the algebraic
failures as a function of test size.

Since the goal is to construct and then test if an evaluation ensemble is
near enough independence to give seemingly correct estimates, a generic
training protocol was applied to the three datasets studied. A training
sample of 600 items was selected and the rest of the dataset was then
held-out to carry out unlabeled evaluations. The rate of failures on the
held-out data is then used as a guide to select test sizes that have low
failure rates as observed empirically.

Independence in the candidate ensembles was maximized by training
each member on features disjoint with those used by the others.
Disjoint partitions of the small 600 training set, each of size 200,
were then used to train each candidate ensemble. A single profiling
run selected 300 disjoint feature partitions to test as the size
of the held out data was changed. Averaging successive profiling runs
then gives an empirical measure of the failure rates as a function of
test size. Each disjoint feature partition was trained and evaluated
ten times.

\begin{figure}
  \centering
  \begin{subfigure}[b]{0.45\textwidth}
         \centering
         \includegraphics[width=\textwidth]{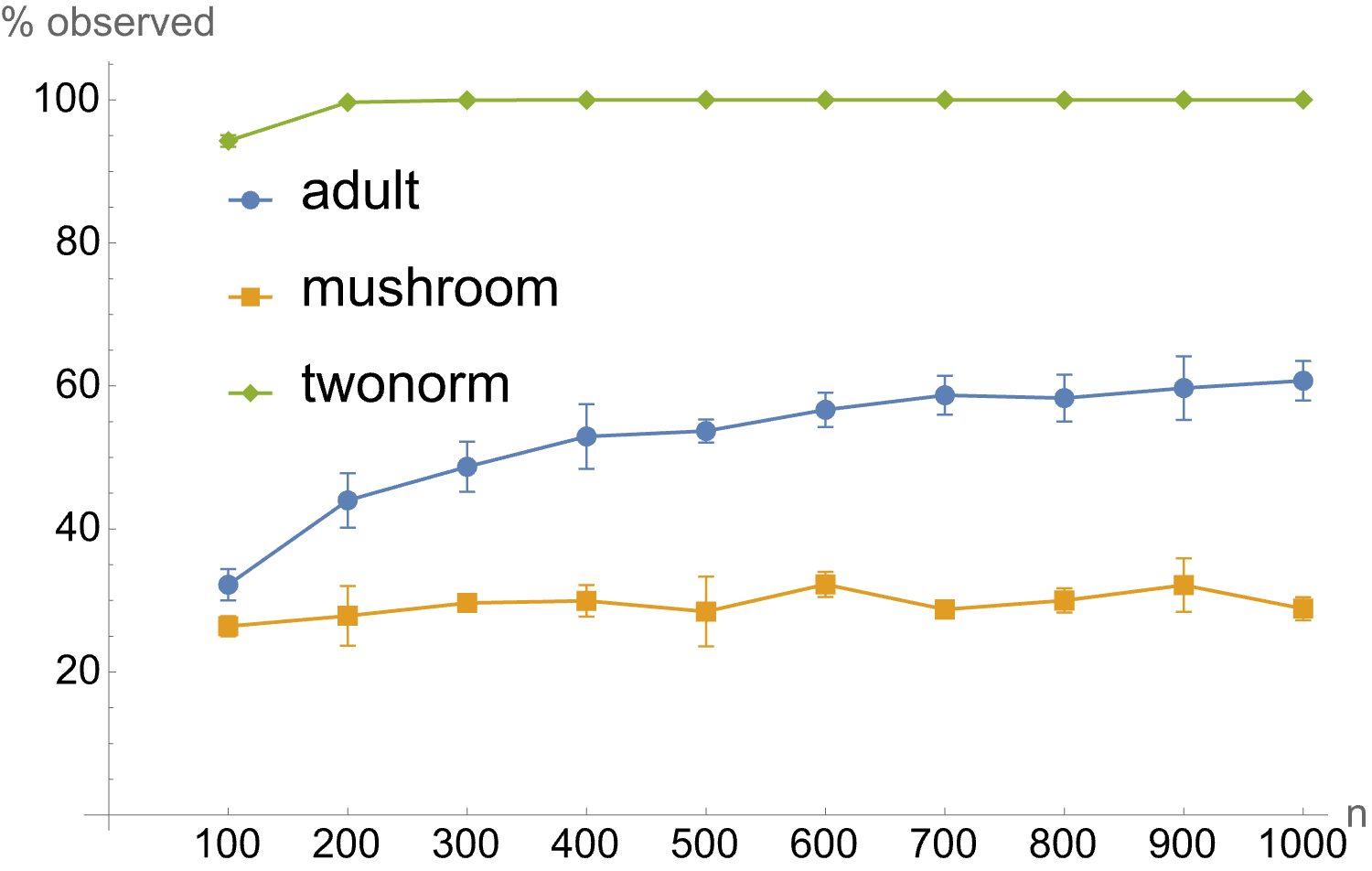}
         \caption{Percentage of seemingly correct feature partitions by test size.}
         \label{fig:a}
  \end{subfigure}
  \hfil
  \begin{subfigure}[b]{0.45\textwidth}
         \centering
         \includegraphics[width=\textwidth]{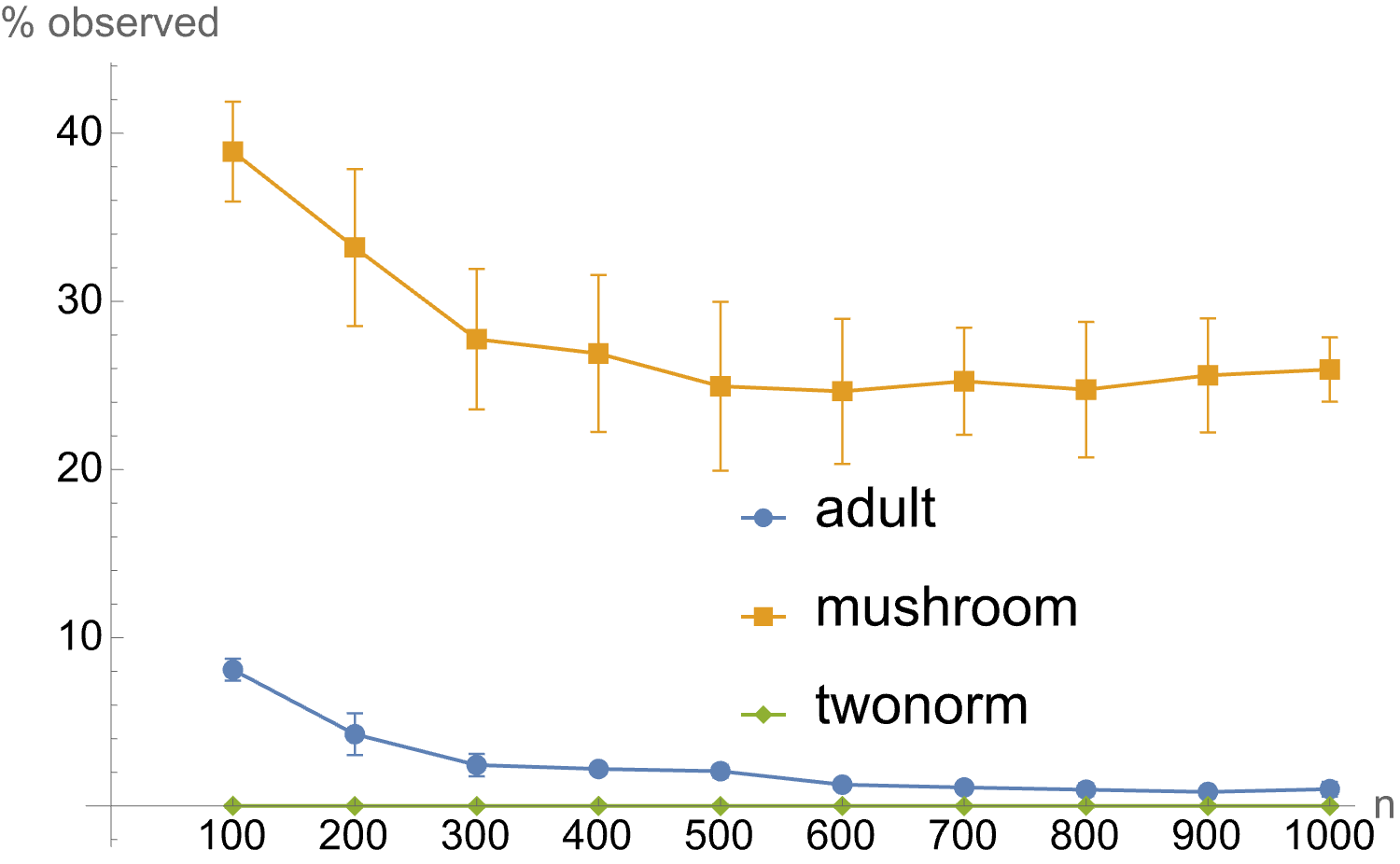}
         \caption{Percentage of feature partitions that never had an independent
         model solution.}
         \label{fig:b}
  \end{subfigure}
  \caption{Failure rates for candidate evaluation ensembles 
  constructed from disjoint partitions of the features.
  }
\end{figure}

Figure~2 shows profiles of algebraic failures for the \texttt{adult}, 
\texttt{mushroom}, and \texttt{twonorm} dataset experiments. Figure~\ref{fig:a} plots
the percentage of feature partitions that resulted in evaluation ensembles returning
seemingly correct estimates. The \texttt{twonorm} and
\texttt{adult} experiments suggest that nearly independent ensembles
will be easier for them than in \texttt{mushroom}. Figure~\ref{fig:b} plots the
percentage of feature partitions that never produced data sketches
explainable by the independent assumption. This plot also suggests that
\texttt{mushroom} good evaluation ensembles will be harder to find.

\section{A containing variety for arbitrarily correlated binary classifiers}

The rejection of evaluation ensembles that have algebraic failures is no
guarantee that the remaining ones will return somewhat accurate evaluation estimates.
A different criteria has to be found to identify those that are close to
independence. We can do this by considering the evaluation ideal of arbitrarily
correlated classifiers.

Theorem 3 in the Supplement shows how a complete polynomial representation for
their data sketch is possible when we include variables for each of
the correlation statistics. Its corresponding evaluation variety remains an open
problem. But another variety that contains it can be defined. And most importantly,
it can be constructed without knowledge of the correlation statistics. Theorem 4
in the Supplement details how the basic evaluation statistics can be disentangled
from the correlation ones by finding a suitable Gr\"{o}bner basis for the generating
set. The polynomials of the disentangled set have the forms,
\begin{gather}
    \prva \, (\prsa{i} - f_{i,\lbla}) = (1-\prva) \, (\prsb{i} - f_{i,\lblb}) \\
     (\prsa{i} - f_{i,\lbla}) \, (\prsb{j} - f_{j,\lblb}) =  (\prsb{i} - f_{i,\lblb}) \, (\prsa{j} - f_{j,\lbla}).
\end{gather}
Since this generating set is a subset of the complete generating set for correlated classifiers,
it is guaranteed to contain their evaluation variety, which, in turn, must contain the
true evaluation point.

\begin{figure}
  \label{fig:max-corr-vs-distance}
  \centering
  \begin{subfigure}[b]{0.45\textwidth}
         \centering
         \includegraphics[width=\textwidth]{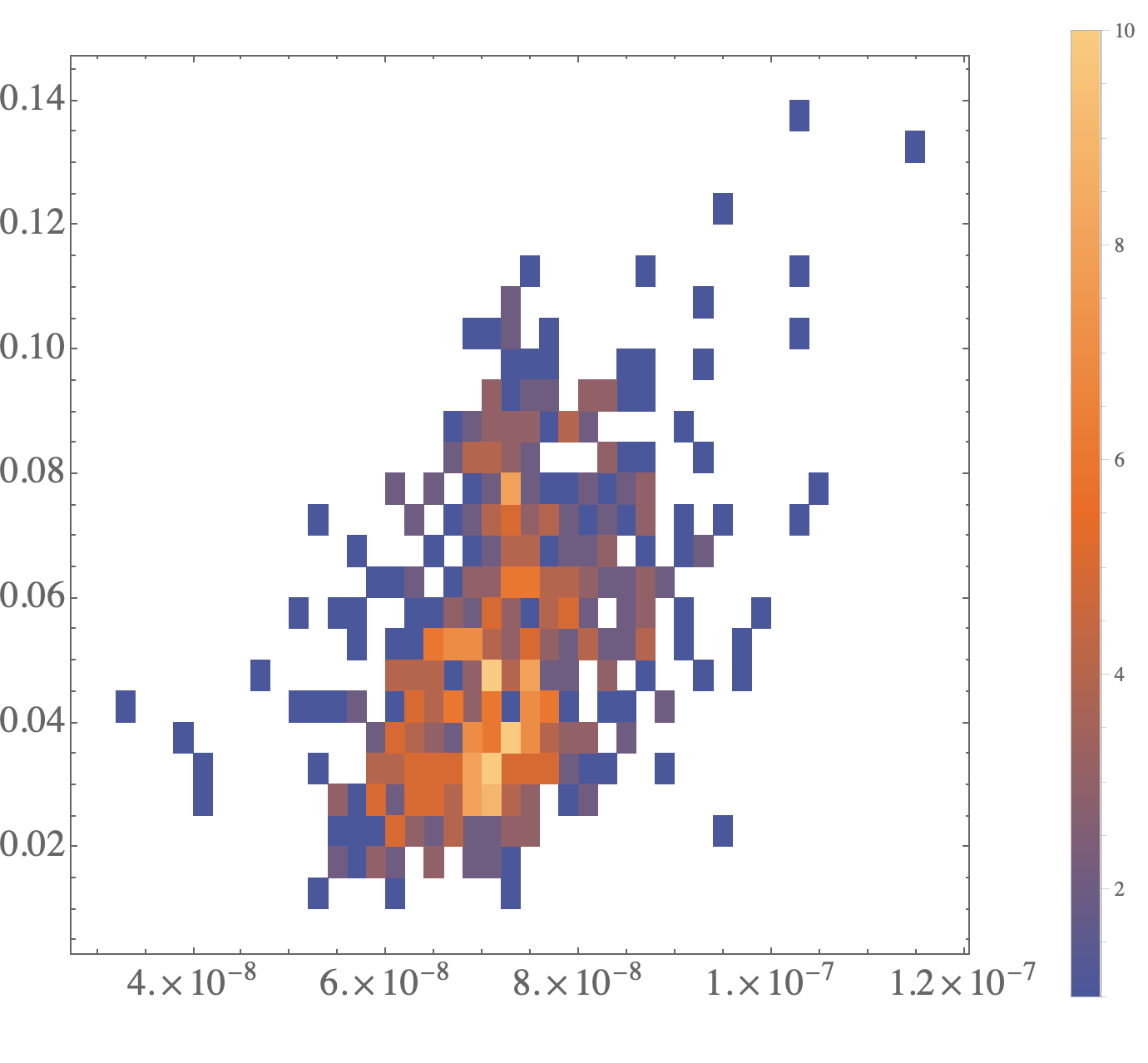}
         \caption{The \texttt{adult} dataset.}
         \label{fig3:a}
  \end{subfigure}
  \hfil
  \begin{subfigure}[b]{0.45\textwidth}
         \centering
         \includegraphics[width=\textwidth]{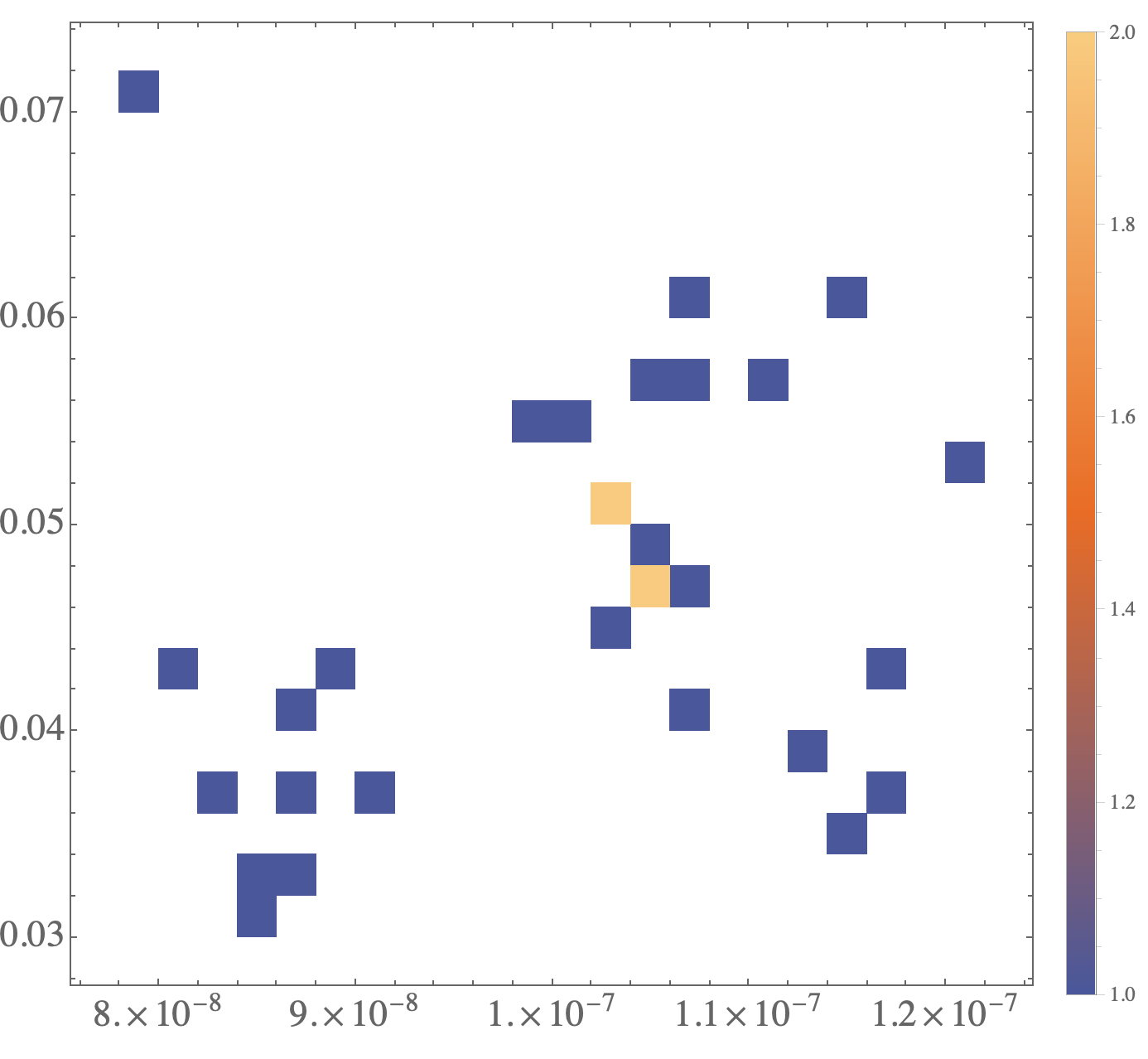}
         \caption{The \texttt{mushroom} dataset.}
         \label{fig3:b}
  \end{subfigure}
  \caption{Maximum pair correlation in an evaluation ensemble versus 
  the distance of the independent evaluator estimate it produces
  from the containing variety.}
\end{figure}

By self-consistency, if the evaluation ensemble was truly independent, it would
be on this 4-dimensional surface. The second set of experiments looked at the hypothesis that
data sketches from correlated classifiers would return independent evaluator
estimates whose distance from the containing variety was related to their
unknown amount of correlation.
Figures 3a and 3b show the observed relation between the distance to the containing
variety and an ensemble's maximum absolute pair correlation for the \texttt{adult} and \texttt{mushroom}
experiments. As in the first experiments, the training data was 200 per label items. 
But the evaluation was carried out on held-out data with 2000 per label items. 
As was expected by the test size profiling runs,
the \texttt{mushroom} seemingly correct evaluations were harder to find. The trend
in these plots is suggestive but not conclusive.

The Supplement contains some of the evaluations from the least-distance evaluation
ensembles. The evaluations on \texttt{twonorm} perform best, perhaps because
that dataset is synthetic. But challenges remain when handling correlated
ensembles. Perhaps these will be resolved with further work. This may be possible
because we have a complete representation for correlated classifiers. Using
that representation one can expand the independent evaluator estimates as
Taylor series on the unknown correlations. The linear term in \corrtwo{i}{j}{\ell}
has the inverse,
\begin{equation}
    1/(\prs{k}{\ell} - f_{k,\ell}).
\end{equation}
Consequently, independent evaluator estimates becomes worse the closer one is to
the ``blindspots'' in evaluation space at $f_{i,\ell}.$ A look at the Gr\"{o}bner
basis for correlated classifiers shows how the blindspots shunt off the correlation
variables by eliminating them from the basis. An evaluator whose evaluation statistics
lie at the blindspots is thus unable to capture correlation effects - its sketch is
explainable by an independent ensemble hypothesis that is not correct. This extreme
case happens at a finite number of points, so its occurrence would be correspondingly
rare. But as the Taylor expansion shows, it can affect the quality of the
independent evaluator estimate severely if the true evaluation point lies near them.

\section{Advantages and disadvantages of algebraic stream evaluation}
The main advantage of algebraic evaluation is that it bypasses the
representation and out-of-distribution problems in ML. Its focus is on
estimating sample statistics with no concern for inferring models
of the phenomena being classified or how the classifiers do it. There
are no unknown unknowns in algebraic evaluation.

But algebraic evaluation cannot resolve the principal/agent monitoring paradox,
only ameliorate it. Its batch approach only estimates average performance
on a test. This may not be sufficient to handle anomalies or identify 
important subsets of the test where the classifiers perform much worse.
In addition, sample statistics are not enough to identify the causes
of poor performance or predict performance in the future. These are important
considerations in settings one bothers to monitor with evaluation ensembles.
Algebraic evaluators should be used in conjunction with other evaluation
methods, such as the ones discussed in Section~\ref{sec:previous-work}, that
do encode more information about the application context.

Finally, all evaluation methods on unlabeled data are
ambiguous. This is seen here by the two-point variety associated
with independent classifiers. Additional assumptions about the evaluation
must be made to `decode` the true evaluation point. For example, in
contexts where the prevalences are not expected to vary greatly their
known value can be used to select to correct point. Such is the case
in the \texttt{adult} dataset where the rare label corresponds to tax
record features for people earning more than 50K US dollars annually.
Fewer higher income records is a reasonable assumption for future
random samples of US tax records. Conversely, if one could have high
assurances of the quality of the classifiers and then use them to
select the one point that aligns with it. In that case, stream evaluation
is being used to monitor the environment and not the classifiers.

\section{Broader Impacts}

Evaluation on unlabeled data is a perennial problem in ML. As this conference
and others discuss the impact AI agents have on our safety and society, it becomes
necessary to have safeguards that can protect us from their decisions. The
framework proposed here should have a positive impact across multiple
application areas for ML since it is based on generic considerations.

\begin{ack}
Simeon Simeonov, and Real Chemistry generously supported the writing of this paper.
\end{ack}

\bibliographystyle{unsrtnat}
\bibliography{streamalgeval}

\end{document}


\maketitle

\section{Introduction}

The methodology of using algebraic geometry for finite test evaluation
is quite involved and uses mathematics not commonly mentioned in the
ML literature. This supplement will try to gently guide the reader to
understanding these tools as it provides proofs for the theorem's
mentioned in the paper. It then does a comparison with Theorem 1 from
Jaffe et al.\ for independent classifiers that should be considered
the probabilistic counterpart of the algebraic approach used here.
Finally, it closes with a detailed discussion of the main experiments
in the paper as well as some additional ones.

\subsection{The general idea of algebraic evaluation}

The paper focuses on binary classification but the methodology described
is applicable to any number of labels. In addition, it can be used to
study decision data sketches for events besides those at the per-item level.
This more general framework can be stated as follows -
\begin{enumerate}
    \item Define the decision event data sketch for the classification stream.
    \item Equate each possible decision event to a sum over the true labels.
    \item Use the true label indicators to construct exact polynomials describing
    a label's contribution to the observed frequency.
    \item Compute the set of points in evaluation space, the evaluation variety,
    that can explain the observed data sketch.
\end{enumerate}
These steps are possible for finite tests because one can always find finite
moment expansions for any sample statistic. In essence, we are guaranteed to be
able to formulate and prove that a particular polynomial representation can
explain all observable data sketches. These polynomial representations may
be quite involved but modern computer algebraic systems have no difficulty
handling their construction. The last step, finding its corresponding variety 
is the hard part.

The purpose of an evaluation is to get a `grade' for the ensemble members -
an actual number. We will abuse notation by using the same symbols to
express the value in an actual test versus the variable used to carry out
algebraic formulations. Thus, the prevalence of label \lbla , \prva, refers both
to its actual value in a test and the variable that defines one of the dimensions
in \emph{evaluation space} - the space defined by variables associated to each
sample statistic.

\subsection{The postulate of true or ground truth labels}

All the items in the stream have a true label. This ground
truth is expressed by the ground truth indicator functions, \indgt{s}{\ell}.

\begin{defn}
For each item, $s$, in the stream there is a ground truth label
indicator function,
\indgt{s}{\ell}, given by,
\begin{equation}
    \indgt{s}{\ell}=\begin{cases}
        1 & \text{if $\ell = \ell_{\text{true}}^{(s)}$}\\
        0 & \text{otherwise}
    \end{cases}
\end{equation}
\end{defn}
The existence of true labels for any one item is then expressed mathematically by,
\begin{equation}
    \label{eq:sumone}
    \sum_{\ell_i \in \{\alpha, \beta, \gamma, \ldots \}} \indgt{s}{\ell_i} = 1.
\end{equation}
For binary classification there are only two terms in this equation. We could 
thus choose to represent events by polynomials that use variables expressing
the frequency of getting a label correct versus not. That is, there is only
one way to be wrong. But for three or more labels there is more than one way
to be wrong. In such cases, it would be more natural 
(i.e.\ result in more symmetric polynomials) if all decision
events are expressed in terms of the frequencies of labels wrong. The paper
chose to describe binary classification events in terms of variables
quantifying the frequency of correct label decisions.

\subsection{Definitions of the data sketch and its associated sample statistics}

The black-box approach to noisy judges taken here is that the only information
available for evaluation are observations of their decisions. In the classification
task that means that for classifier $i$ in the ensemble we can observe its decisions on
the stream items,
\begin{equation}
    \{ \lis{i}{s} \}_{s=1}^{n},
\end{equation}
where $n$ is the number stream items observed so far.

\begin{defn}
    A \emph{per-item ensemble decision event} for item $s$ is the ordered tuple 
    (\lis{1}{s}, \lis{2}{s}, \ldots).
\end{defn}
From these events we can construct a corresponding data sketch that forgets
any information about decisions across items and just tallies the per-item
decision events.

\begin{defn}
    The \emph{per-item decision data sketch} for an ensemble of $C$ classifiers is
    given by the integer counters for all possible per-item decision events. For $L$ possible
    labels, $L^C$, counters are needed.
\end{defn}

\subsubsection{The label prevalences}

The prevalences of the labels in the stream are integer ratios defined using
the true label indicator functions.
\begin{defn}
    The true prevalence of a label $\ell$, \prv{\ell}, in the observed stream items
    is given by,
    \begin{equation}
        \prv{\ell} =  \frac{1}{n} \sum_{s=1}^{n} \indgt{s}{\ell}
    \end{equation}
\end{defn}
It follows from the postulate of ground truth labels that,
\begin{equation}
    \sum_{\ell \in \{\alpha, \beta, \gamma, ...\}} \prv{\ell} = 1
\end{equation}
This equation is not mentioned again here because for binary classification
we can make it disappear by focusing on just one of the two label prevalences.
In general, this would not be possible and this equation would be
part of the polynomial set that defines the evaluation ideal.

Note that when one wants to evaluate performance across stream items, there would
be prevalences for each of the true label tuples possible. For example, to
evaluate accuracies on two consecutive stream items, four prevalences would be
required during binary classification.

\subsubsection{Classifier label accuracies}

The performance of an ensemble classifier for $L$ labels requires as many
sample statistics. But by Equation~\ref{eq:sumone}, one of these could be
expressed in terms of the others. In binary classification it is sufficient
to use the accuracies on each label.
\begin{defn}
    The \lbla \, and \lblb \, accuracies for classifier $i$ are given by,
    \begin{align}
        \prsa{i} &= \frac{1}{n_\alpha} \sum_{\indgt{s}{\alpha} = 1} \indgt{s}{\lis{i}{s}} \\
        \prsb{i} &= \frac{1}{n_\beta} \sum_{\indgt{s}{\beta} = 1} \indgt{s}{\lis{i}{s}} 
    \end{align}
\end{defn}

\subsubsection{Classifier decision correlations}

The decision correlations during a finite test are defined as products of terms
in the form,
\begin{equation}
    \indgt{s}{\lis{i}{s}} - \prs{i}{\ell}.
\end{equation}
For ensembles of three classifiers we only need two and three way correlation variables
for each label. These are defined as follows.
\begin{defn}
    The \emph{2-way or pair decision correlation} for classifiers $i$ and $j$ on label
    $\ell$ is given
    by,
    \begin{equation}
        \corrtwo{i}{j}{\ell} = \frac{1}{n_\ell} \suml (\indgt{s}{\lis{i}{s}} - \prs{i}{\ell})
        (\indgt{s}{\lis{j}{s}} - \prs{j}{\ell}).
    \end{equation}
    The \emph{3-way decision correlation} for classifiers $i$, $j$, and $k$ on label $\ell$
    is expressed as,
    \begin{equation}
        \corrthree{i}{j}{k}{\ell} = \frac{1}{n_\ell} \suml (\indgt{s}{\lis{i}{s}} - \prs{i}{\ell})
        (\indgt{s}{\lis{j}{s}} - \prs{j}{\ell}) (\indgt{s}{\lis{k}{s}} - \prs{k}{\ell}).
    \end{equation}
\end{defn}

\subsubsection{Finite moment expansions}

Other than carefully defining how these variables are defined in terms of the ground truth indicator
functions, there is nothing novel or unusual here. Expansions of sample statistics in terms of
data moments is well-known. For three classifiers we require the first moment variables (the label
accuracies), the second moment variables (the pair correlations) and the third moment variables
(the 3-way correlations).

\section{Theorem 1: the polynomial generating
set for independent classifiers}

It is possible to define mathematical objects that do not exist.
Hence, formal mathematical treatments start by establishing that
they exist - existence theorems. Another class of theorems are
about completeness of representations - that we have a way of describing
all possible objects. For example, the completeness of Fourier series
in describing all possible piecewise continuous functions. Theorem
1 is a combination of these two types of theorems. It proves that
the decision data sketch of independent classifiers is exactly described
by polynomials using only the variables in the \emph{basic set} of statistics:
\prva, \setpria, and \setprib. But because the proof is constructive and starts
from the true evaluation point, it also proves that the constructed polynomials
do define a variety that is non-trivial (not the empty set).

\begin{thm}
    Each of the decision event frequencies, \freqthree{\ell_1}{\ell_2}{\ell_3},
    derived from the per-item data sketch of three independent classifiers is given by
    a polynomial in the basic set of statistics. They are the \emph{generating set}
    of the evaluation ideal for independent classifiers,
    \begin{flalign}
  \freqthree{\alpha}{\alpha}{\alpha}  &=  \prva \prs{1}{\alpha} \prs{2}{\alpha} \prs{3}{\alpha}& + & \; (1 - \prva) (1 - \prs{1}{\beta}) (1 - \prs{2}{\beta}) (1 - \prs{3}{\beta})\\
  \freqthree{\alpha}{\alpha}{\beta}  &=  \prva  \prs{1}{\alpha} \prs{2}{\alpha} (1 - \prs{3}{\alpha}) & + & \;(1 - \prva) (1 - \prs{1}{\beta}) (1 - \prs{2}{\beta}) \prs{3}{\beta}\\
  \freqthree{\alpha}{\beta}{\alpha}  &=  \prva  \prs{1}{\alpha}  (1 - \prs{2}{\alpha})  \prs{3}{\alpha} & + & \;(1 - \prva) (1 - \prs{1}{\beta}) \prs{2}{\beta} (1 - \prs{3}{\beta})\\
  \freqthree{\beta}{\alpha}{\alpha}  &=  \prva  (1 - \prs{1}{\alpha}) \prs{2}{\alpha} \prs{3}{\alpha} & + & \;(1 - \prva) \prs{1}{\beta} (1 - \prs{2}{\beta}) (1 - \prs{3}{\beta})\\
  \freqthree{\beta}{\beta}{\alpha}  &=  \prva  (1 - \prs{1}{\alpha}) (1 - \prs{2}{\alpha}) \prs{3}{\alpha} & + & \;(1 - \prva) \prs{1}{\beta} \prs{2}{\beta} (1 - \prs{3}{\beta})\\
  \freqthree{\beta}{\alpha}{\beta}  &=  \prva  (1 - \prs{1}{\alpha})  \, \prs{2}{\alpha} \, (1 - \prs{3}{\alpha}) & + & \; (1 - \prva) \prs{1}{\beta} (1 - \prs{2}{\beta}) \prs{3}{\beta}\\
  \freqthree{\alpha}{\beta}{\beta}  &= \prva  \prs{1}{\alpha}  (1 - \prs{2}{\alpha})  (1 - \prs{3}{\alpha}) & + & \;(1 - \prva) (1 - \prs{1}{\beta}) \prs{2}{\beta} \prs{3}{\beta}\\
  \freqthree{\beta}{\beta}{\beta}  &= \prva  (1 - \prs{1}{\alpha})  (1 - \prs{2}{\alpha}) (1 - \prs{3}{\alpha}) & + & \; (1 - \prva) \prs{1}{\beta} \prs{2}{\beta} \prs{3}{\beta}
\end{flalign}
  This generating set defines a non-empty evaluation variety that contains the true evaluation point.
\end{thm}
 \begin{proof}
 By the existence of true labels, it follows that any decision event by the classifiers has a count equal
 to the sum of times the true labels were \lbla \, plus those were the true labels were \lblb,
     \begin{equation}
    n_{\ell_1, \ell_2, \ell_3} = \#(\ell_1, \ell_2, \ell_3 \svbar \lbla) +
    \#(\ell_1, \ell_2, \ell_3 \svbar \lblb).
    \label{eq:label-decomposition}
\end{equation}
Dividing this equation by $n$, the number of items classified and then multiplying
each label term by unity in the form $n_\ell/n_\ell$, this becomes
\begin{align}
    \freqthree{\ell_1}{\ell_2}{\ell_3} &= \frac{n_\alpha}{n} (\frac{1}{n_\alpha} \#(\ell_1, \ell_2, \ell_3 \svbar \lbla))
    + \frac{n_\beta}{n} (\frac{1}{n_\beta} \#(\ell_1, \ell_2, \ell_3 \svbar \lblb)) \\
                                       &= \prva (\frac{1}{n_\alpha} \#(\ell_1, \ell_2, \ell_3 \svbar \lbla)) +
                                       \prvb (\frac{1}{n_\beta} \#(\ell_1, \ell_2, \ell_3 \svbar \lblb)) \\
                                       &= \prva (\frac{1}{n_\alpha} \#(\ell_1, \ell_2, \ell_3 \svbar \lbla)) +
                                       (1-\prva) (\frac{1}{n_\beta} \#(\ell_1, \ell_2, \ell_3 \svbar \lblb))
\end{align}
The construction of the generating set then proceeds by reformulating the number of times a decision event
occurred given the true label in terms of the label accuracies of the classifiers. This is tedious but
straightforward. 

Consider the decision event (\lbla, \lblb, \lbla). When the true label is \lbla \, the number of times this
event occurred is equal to,
\begin{equation}
    \label{eq:composite-a}
    \#(\lbla, \lblb, \lbla \svbar \lbla) = 
    \suma \indgt{s}{\lis{1}{s}} (1 - \indgt{s}{\lis{2}{s}}) \indgt{s}{\lis{3}{s}}.
\end{equation}
Correspondingly, the composite indicator function for (\lbla, \lblb, \lbla) events when the true label is \lblb \, is given by,
\begin{equation}
    \label{eq:composite-b}
    \#(\lbla, \lblb, \lbla \svbar \lblb) = 
    \suma (1 - \indgt{s}{\lis{1}{s}})  \indgt{s}{\lis{2}{s}} (1 - \indgt{s}{\lis{3}{s}}).
\end{equation}
The proof now hinges on whether we can write averages of products of the true indicator
functions as products of their averages when the classifiers are independent. 

Every decision event composite indicator function like those in Equations~\ref{eq:composite-a}
and \ref{eq:composite-b} will contain at most pair products,
\begin{equation}
    \indgt{s}{\lis{i}{s}} \indgt{s}{\lis{j}{s}},
\end{equation}
and triple products,
\begin{equation}
    \indgt{s}{\lis{i}{s}} \indgt{s}{\lis{j}{s}} \indgt{s}{\lis{j}{s}}.
\end{equation}
Let us look at the pair product using the definition for the pair correlation
variables and go through simplifications that should be familiar,
\begin{align}
    \corrtwo{i}{j}{\ell} & = \frac{1}{n_\ell} \suml (\indgt{s}{\lis{i}{s}} - \prs{i}{\ell})
        (\indgt{s}{\lis{j}{s}} - \prs{j}{\ell}) \\
                         & \label{eq:corrtwo} = \left( \frac{1}{n_\ell} \suml \indgt{s}{\lis{i}{s}} \indgt{s}{\lis{i}{s}} \right)
                         - \prs{i}{\ell} \prs{j}{\ell}.
\end{align}
Setting the pair correlation variables to zero then guarantees that averages of products are products of
averages. In addition, it forms part of the definition of what `independence' means when talking about
sample statistics. Independent classifiers must have all pair correlation values, \corrtwo{i}{j}{\ell},
set to zero.

The triple product is more involved but follows accordingly from the definition of the 3-way correlation
values and the pair ones,
\begin{align}
    \corrthree{i}{j}{k}{\ell}  = & \quad \frac{1}{n_\ell} 
                                   \suml (\indgt{s}{\lis{i}{s}} - \prs{i}{\ell})
                                         (\indgt{s}{\lis{j}{s}} - \prs{j}{\ell}) 
                                         (\indgt{s}{\lis{k}{s}} - \prs{k}{\ell})\\
                               \begin{split}
                               = & \quad \left( \frac{1}{n_\ell} \suml \indgt{s}{\lis{i}{s}} \indgt{s}{\lis{j}{s}} \indgt{s}{\lis{k}{s}} \right) - 
                                   \prs{i}{\ell} \left( \frac{1}{n_\ell} \suml  \indgt{s}{\lis{j}{s}} 
                                                                                \indgt{s}{\lis{k}{s}} \right) - \\
                                 &\prs{j}{\ell} \left( \frac{1}{n_\ell} \suml  \indgt{s}{\lis{i}{s}} 
                                                                                \indgt{s}{\lis{k}{s}} \right) - 
                                   \prs{k}{\ell} \left( \frac{1}{n_\ell} \suml  \indgt{s}{\lis{i}{s}} 
                                                                                \indgt{s}{\lis{j}{s}} \right) + \\
                                    &2 \prs{i}{\ell} \prs{j}{\ell} \prs{k}{\ell}
                               \end{split}.
\end{align}
Repeatedly invoking Equation~\ref{eq:corrtwo}, the final expression for the 3-way correlation values is,
\begin{multline}
    \corrthree{i}{j}{k}{\ell} = \left( \frac{1}{n_\ell} \suml \indgt{s}{\lis{i}{s}} \indgt{s}{\lis{j}{s}} \indgt{s}{\lis{k}{s}} \right) + \prs{i}{\ell} \corrtwo{j}{k}{\ell} + \prs{j}{\ell} \corrtwo{i}{k}{\ell} \\
    + \prs{k}{\ell} \corrtwo{i}{j}{\ell} - \prs{i}{\ell} \prs{j}{\ell} \prs{k}{\ell}
\end{multline}
This defines the final condition for the sample independence of three classifiers - both 
\corrthree{i}{j}{k}{\lbla} and \corrthree{i}{j}{k}{\lblb} are equal to zero. This completes
the proof that the generating polynomial set in the theorem is sufficient to explain all decision
data sketches by independent classifiers using only polynomials of the basic set of evaluation
variables.

The proof quickly concludes by observing that the constructive part of the proof means that we have
at least one point in the evaluation space defined by the basic set that satisfies all these equations
when we go from thinking about them as actual values to treating them as variables. The set of points
that satisfy the equations for a polynomial set is called the \emph{variety} or as is being called here
- the \emph{evaluation variety}. It is non-empty and it contains the true evaluation point.
\end{proof}

\section{Theorem 2: the evaluation variety
for independent classifiers}

Theorem 1 constructed the polynomial generating set for independent classifiers and showed that the
true evaluation point is contained in its variety. Theorem 2 now considers the question of how many
points, besides the true evaluation point, could also explain an observed decision data sketch from
independent classifiers.

\begin{thm}
    The evaluation variety of the independent classifiers generating set contains exactly
    two points, one of which is the true evaluation point.
\end{thm}
\begin{proof}
    Solving the generating set for independent classifiers is quite involved algebraically.
    The accompanying Mathematica notebook foo details the calculations. Here we describe
    the general strategy of the proof and add more explanations for the terms involved
    in the independent evaluator's expression for \prva that appears in Table 1 in the paper.
    
    A strategy for solving multi-variable polynomial system
    is to obtain an equivalent representation of the polynomials that create an elimination
    `ladder'. At the bottom of the ladder is a single variable polynomial that, by the fundamental
    theorem of algebra, has as many roots (counting multiplicity) as the order of the polynomial.
    One then climbs up the ladder by finding polynomials that involve the solved variable and
    one more variable. In this manner, one systematically solves for the unknown values for the
    variables satisfying the polynomial system. This alternative representation is called the
    \emph{elimination ideal}. It can be obtained by solving for the Gr\"{o}bner basis of
    the generating set. The term `basis' is somewhat unfortunate since it may mislead the reader
    into thinking that a polynomial basis is like a vector basis. It is not. For example, the
    number of polynomials in a given basis for an evaluation ideal need not be the same as that
    in another basis. There is no concept of dimension when talking about evaluation ideals.

    Buchberger's algorithm is a generic algorithm for finding the many Gr\"{o}bner bases one
    may care to obtain. The Gr\"{o}bner basis used in the notebook computations for this proof
    is also an elimination ideal. It contains at the bottom a quadratic polynomial in \prva of
    the form,
    \begin{equation}
    \label{eq:prev-quadratic}
    a(\ldots) \prva^2 + b(\ldots) \prva + c(\ldots) = 0.
    \end{equation}
    By the fundamental theorem of algebra, this has two roots. And by Theorem 1 we know that
    one must be the true prevalence for the \lbla \, items. Because the elimination ideal also
    contains linear equations relating a single \prs{i}{\ell} variable with \prva, it also follows
    that each root of the \prva polynomial is associated with a single value for all the \prs{i}{\ell}
    variables. This proves that the evaluation variety for independent classifiers contains exactly
    two points in evaluation space.

    The two points are related. Given a point that solves the independent generating set, the other
    one is given by the transformations
    \begin{align}
        \prva & \rightarrow (1-\prva)\\
        \prs{i}{\lbla} & \rightarrow (1 - \prs{i}{\lblb})\\
        \prs{i}{\lblb} & \rightarrow (1 - \prs{i}{\lbla})
    \end{align}
\end{proof}
The ambiguity of the exact solution for independent classifiers has been noted before
in the literature. It is inherent to all inverse problems and can be found in many
fields such as tomography and error-correcting codes.

\begin{figure}
  \label{fig:a-coeff}
  \centering
  \includegraphics[width=\textwidth]{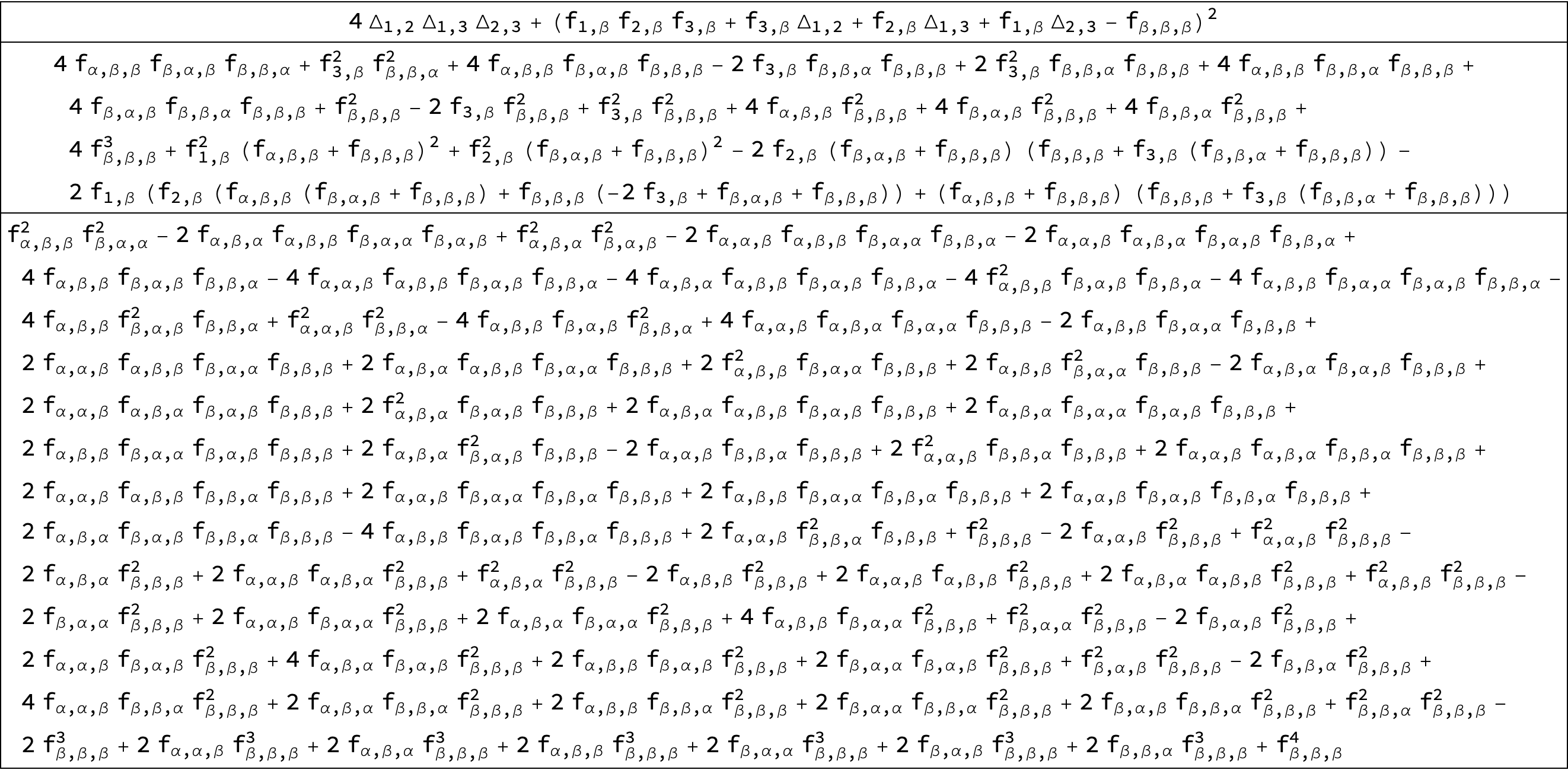}
  \caption{Stages, from the bottom up, of the algebraic simplification of the `a'
  coefficient in the prevalence quadratic (Eq.~\ref{eq:prev-quadratic}). The bottom
  cell shows the expression using the eight data sketch frequencies. The middle
  cell shows the simplification after introducing the \lblb \, label decision frequencies
  for each classifier, $f_{i,\beta}.$ The top cell was obtained by defining new variables,
  $\Delta_{i,j},$ as explained in the main text.
  }
\end{figure}

The raw expression obtained by solving Equation~\ref{eq:prev-quadratic} is unwieldy
and would be impossible to present here. It would also hide some of the structure
that is relevant to understanding the limitations of the formula and of the algebraic
approach in general. In the accompanying Mathematica code (?), the necessary operations
are carried out. For example, Figure~\ref{fig:a-coeff} shows three stages of the simplifications of
the `a' coefficient in the prevalence quadratic (Equation~\ref{eq:prev-quadratic}. The top
cell was obtained by introducing new variables. The $f_{i,\beta}$ variables are the decision
frequencies by the classifier $i$ for the \lblb \, label. These actually define `blind spots'
in the independent evaluator as will be discussed during the proof of Theorem 3. The second
set of new variables, the $\Delta_{i,j}$, are defined as follows,
\begin{equation}
    \Delta_{i,j} = f_{i,j,\lblb} - f_{i,\lblb} f_{j,\lblb}.
\end{equation}
The new decisions frequencies, $f_{i,j,\lblb}$, keep track of how often the pair (i,j)
both voted for label \lblb \, on an item. Interestingly, it can be proven that
\begin{equation}
    f_{i,j,\lblb} - f_{i,\lblb} f_{j,\lblb} = f_{i,j,\lbla} - f_{i,\lbla} f_{j,\lbla}.
\end{equation}
So there is no need to define $\Delta$ variables for each label.

Finally, it should be noted that the 3 independent classifier solution solves the evaluation
problem for any ensemble of three or more of them. This follows from marginalizing
classifiers out. In addition, it can also be shown that 3 is the minimum number needed
for this. The two and one independent classifiers have evaluation variety surfaces,
not points.

\section{Theorem 3: the generating set for
correlated classifiers}

The utility of working with sample statistics comes to the forefront in Theorem 3.
Algebraic evaluation is easy in comparison to predicting. As will be shown by the
theorem, we can write a generating set for data sketches of correlated classifiers
that is complete. This is a universal representation of all data sketches no matter
the setting. There is no theory of the phenomena that the classifiers analyzed in
algebraic evaluation. How can there be when it has no free parameters? Because we
are estimating sample statistics, we can write complete representations. There
are no unknown unknowns in evaluation because of this.

\begin{thm}
There is a polynomial generating set for all per-item data sketches by correlated
classifiers. It requires the basic set of evaluation statistics \prva, \setprian,
\setpribn. In addition, it uses n choose two pair correlation variables per label,
n choose three 3-way correlation variables and so on until it terminates at 1 n-way
correlation variable. It defines an evaluation variety that contains at least one
point - the true evaluation point.
\end{thm}
\begin{proof}[Sketch of the proof]
The proof is constructive as in Theorem 1. We are starting at the true evaluation
point. This point now exists in a space of dimension,
\begin{equation}
    1 + 2\, n + 2 \, \sum_{m=2}^{n} {n \choose m} = 2^{n+1} -1.
\end{equation}
The frequency of times any decision event occurs given the true label can be written as the average of a composite indicator function as was done in the proof of Theorem 1.
For $n$ classifiers, the largest product term possible in this composite indicator
function would contain $n$ indicator functions. But these can be reformulated in
terms of the basic evaluation statistics, the n-way correlation variable for the label
and averages of products with $n-1$ indicator functions. Continuing in this way,
the products of the $n-1$ can be rewritten in terms of the (n-1)-way correlation
variables and products with $n-2$. This descent ends at $n=1$.

It follows trivially by the constructive nature of the proof that this generating set
defines an evaluation variety that contains at least one point - the true evaluation
point.
\end{proof}

The Mathematica notebook \texttt{GeneratingPolynomialsCorrelatedClassifiers.nb} contains 
explicit formulations of the polynomial generating set for two and three classifier ensembles. 
One practical utility of these general
polynomial sets is that they allow theoretical study of the properties of functions
based on the data sketch frequencies. In particular, one can use these polynomial
expressions when correlation exists to study the behavior of the independent evaluator
estimates. Consider the algebraic expression for \prva \, under the independence assumption.
It is an algebraic function of the data sketch frequencies,
\begin{equation}
    \label{eq:prva}
    \aeprev.
\end{equation}
Self-consistency under the independence assumption means that substituting the independent
generating polynomials for the frequencies into this expression will result in two
solutions,
\begin{equation}
    \prva, 1-\prva.
\end{equation}
But the independent estimate is not correct for correlated classifiers. By using
the generating set for such classifiers, a Taylor expansion on the variables
\corrtwo{i}{j}{\ell} can be done for the \prva \, estimate. This work is not presented
here except to remark that the first order terms have the generic form,
\begin{equation}
    \frac{\corrtwo{i}{j}{\ell}}{\prs{k}{\lbla} + \prs{k}{\lblb} -1}.
\end{equation}
Therefore, the independent evaluator estimates errors grow as one approaches
the line,
\begin{equation}
    \prs{k}{\lbla} + \prs{k}{\lblb} -1,
\end{equation}
in evaluation space. The origin of this `blind spot' and others in algebraic
evaluation will become clearer in the next section where a Gr\"{o}bner basis
for correlated classifiers is discussed.

\section{Theorem 4: A Gr\"{o}bner basis for three correlated classifiers}

The generating set for arbitrarily correlated classifiers quickly overwhelms
current computational commutative algebra platforms such as Mathematica.
They use Buchberger's algorithm, a generic algorithm that is proven to always
terminate in finite time but is known to take exponential time and memory for
some polynomial systems, \cite{Cox}.
Nonetheless, we can get a glimpse of how a general solution looks by solving
for Gr\"{o}bner bases for ensembles of three correlated classifiers.
It is this basis that allows us to define a containing variety, larger than
the evaluation variety, that contains the true evaluation point.

More importantly, the polynomials in the basis are simplified
by expressing them in terms of new variables, $\pi_{i,\ell}$ and
$\gamma_{i,j,\ell}$, that are shifted versions of \prs{i}{\ell} \, and
\corrtwo{i}{j}{\ell} respectively,
\begin{align}
    \pi_{i,\ell} & = \prs{i}{\ell} - f_{i,\ell} \\
    \gamma_{i,j,\ell} & = \corrtwo{i}{j}{\ell} - \Delta_{i,j}.
\end{align}

\begin{thm}
A Gr\"{o}bner basis for the generating set of three correlated classifiers
exists. A subset of the basis consists of polynomials of the forms,
\begin{align}
    \prva \pi_{i,\lbla} & - \prvb \pi_{i,\lblb}\\
    \pi_{i,\lbla} \pi_{j,\lblb} &  - \pi_{i,\lblb} \pi_{j,\lbla}.
\end{align}
These define a \emph{containing variety} guaranteed to contain
the true evaluation point.
\end{thm}
\begin{proof}
The proof is constructive and is given in GroebnerBasis3CorrelatedClassifiers.nb.
The inclusion of the true evaluation point in the containing variety follows from the
general theorem that a subset of generating polynomials must define a variety that
includes the variety of the full set. That is, the set of points that satisfy a subset
of the polynomials has to be equal to or larger than the set of points that satisfy
all the polynomials.
\end{proof}

It is not proven here that the containing variety has dimension $n + 1$ in the
$2n + 1$ space of the basic evaluation statistics of $n$ correlated classifiers.
Since the containing variety does not require any knowledge of the correlation
variables, this dimensionality reduction is universally available and can serve
as a constraint for other methods such as the probabilistic approaches to evaluation.

\section{Theorem 5: Unresolved square roots signal correlation but seemingly
correct estimates are no guarantee of sample independence}

Theorem 5 combines theorems 2 and 3 to prove that unresolved square
roots in the independent evaluators formula for \prva \, can only occur
when the ensemble classifiers are correlated in the evaluation. It
signals, with no false positives, that an observed data sketch was
not produced by an ensemble of independent classifiers. Unfortunately,
the converse is not true precisely because of the blind spots in the
algebraic evaluator discussed in the previous section.

\begin{thm}
    The presence of unresolved square roots in the independent evaluator
    estimate of \prva \, means the classifiers are not sample independent.
    The converse is not true. Integer ratio estimates of \prva \, are
    possible at the algebraic evaluator blind spots.
\end{thm}
\begin{proof}
    An unresolved square root in the \prva estimate cannot be produced by
    sample independent classifiers. This follows immediately from the exact
    solution in Theorem 2. 
    
    That the converse is not true, resolved square
    roots do not mean the classifiers were error independent on the test is
    done by construction. The independent evaluator formula for \prva \,
    (Equation~\ref{eq:prva}) contains a square root denominator of the form,
    \begin{equation}
      \sqrtprevterm.
    \end{equation}
    If any of the delta product terms equals zero, the square root will resolve
    to an integer ratio. We can use the generating set for a pair of correlated
    classifiers to understand when this can happen,
    \begin{align}
    \Delta_{i,j} = & f_{i,j,\lblb} - f_{i,\lblb} f_{j,\lblb} \\
                 = & \left[ \prva  \left( (1 - \prs{i}{\alpha})  (1 - \prs{j}{\alpha}) + \Gamma_{i,j,\lbla} \right) +  
                            \prvb \left(\prs{i}{\beta} \prs{j}{\beta}  + \Gamma_{i,j,\lblb} \right) \right] \\
                   & - \left[ \prva   (1 - \prs{i}{\alpha})   +  \prvb \prs{i}{\beta} \right]
                   \left[ \prva  (1 - \prs{j}{\alpha})  +  \prvb \prs{j}{\beta}  \right] \\
                = & \prva \prvb \left( 1 + \prs{i}{\alpha} + \prs{i}{\beta} \right) \left( 1 + \prs{j}{\alpha} + \prs{j}{\beta} \right) + \left( \prva \Gamma_{i,j,\lbla} + \prvb \Gamma_{i,j,\lblb}\right).
    \end{align}
    These are the blind spot lines discussed before. Setting either to zero eliminates the first term. For
    the second term, depending on the relative sizes of \prva \, and \prvb \, their ratio can be made less
    than 1 and consequently there is another line defined by,
    \begin{equation}
        \prva \Gamma_{i,j,\lbla} + \prvb \Gamma_{i,j,\lblb} = 0,
    \end{equation}
    such that given one correlation value between -1 and 1, we can find another correlation in the same range that
    zeros this term out also. We conclude that there are correlated classifier ensembles that give seemingly
    correct \prva \, estimates.
\end{proof}

\section{Detailed comparison with \cite{Jaffe2015}}

The natural counterpart to the exact solution provided in Theorem 2 is Theorem 1
in \cite{Jaffe2015}. There are various reasons for this. As mentioned in the
Previous Work section, their paper follows the spectral approach to evaluation
started by \cite{Parisi1253}. As such, there are no hyperparameters as in Bayesian
approaches to evaluation. But like all probability approaches for evaluation, their
work is concerned with inferring unknown distributions in the limit of infinite
sample sizes.

The key to understanding the difference between any probabilistic approach and
the one taken here is that they are discussing different notions of independence.
The distributional independence assumption in \cite{Jaffe2015} is not equivalent
to the sample independence assumption used in this paper. This is illustrated by
the considering how to construct data sketches of independent classifiers
under either approach. In the algebraic approach taken here you could use
the generating set of polynomials in Theorem 1 to do this quickly. Set the prevalence
and label accuracies to desired integer ratios, plug them into the polynomials and
compute the resulting data sketch. The computed decision event frequencies can then
be used to compute the minimum test size that would have given that data sketch -
the GCD of the frequencies. We can then just do random shuffles of these decision
events to create a very large number of stream simulations, all of which have
zero correlation on the test and have exactly the same label prevalences and 
classifier accuracies.
Consider now trying to get the same data sketch by simulating an i.i.d.\ process
using the same settings as was done with just the algebra of the generating set.
The resulting data sketch will almost certainly not be one that has zero
sample correlations.

Their solution to distributionally independent classifiers is based on three
assumptions. They are,
\begin{enumerate}
    \item The stream items (`instances' in their paper) are assumed to be i.i.d.
    realizations of the marginal for them, $p_{X}(x)$, for an unknown joint distribution
    of the items and the classifiers decisions.
    \item The classifiers are conditionally independent for any pair - their joint
    decision distribution is the product of their individual distributions.
    \item More than half the classifiers are assumed to have label accuracies that sum
    to more than 1.
\end{enumerate}
Assumption 1 is irrelevant in algebraic evaluation. It does not matter if the stream items
are not i.i.d.\ under any distribution. Sample statistics about per-item events have nothing
to do with sample statistics across items. Assumption 2 is the distributional definition of
independence. Assumption 3, as they themselves mention, is strictly speaking not necessary.
It relates to `decoding` the true evaluation point. The point of view of this paper is that
this is a separate concern, as discussed in Section 6 of the paper. Identifying the true
point will always require additional side-information. Assumption 3 is akin to error-correcting
codes deciding that the least bit flips solution is the correct one when an error is detected
and corrected. Relying on prevalence knowledge is another acceptable way to decode the true
evaluation point.

The terminology of \cite{Jaffe2015} uses \emph{specificity} and \emph{sensitivity} to
describe what are here called the label accuracies. Interestingly, later in the paper, when
they consider 3 or more label classification, they revert to using the terminology of confusion
matrices - the one essentially used here. To continue the comparison with their work, we arbitrarily
decide that what they call label "1" is \lbla \, and their label "-1" is \lblb. Under that identification.
the mapping to sensitivity and specificity variables is,
\begin{align}
  \psi_i & \rightarrow \prsa{i}\\
  \eta_i & \rightarrow \prsb{i}
\end{align}
But care must be taken not to equate the two. The specificity and sensitivity are referring to unknown, 
discrete distributions. The label accuracies in this paper refer to sample averages, not distributions.
The predictions of the classifiers are represented by functions, $f_i(X)$, that yield `1' and `-1'. This
entanglement of labels and values then leads them to consider the infinite sample size quantities,
\begin{align}
    \mu_i & = \mathbb{E} \left[ f_i(X) \right] \\
    R & = \mathbb{E}\left[(f_i(X) - \mu_i)(f_j(X) - \mu_j)\right]
\end{align}
The crux of the spectral approach is that the off-diagonal elements of the matrix $R$ are identical
to a rank-one matrix $\textbf{v}\textbf{v}^T$ where the vector $\textbf{v}$ encodes the class imbalance in
the labels and the average of the label accuracies.
But these quantities are not known for any finite sample so they must consider their finite
test averages and consequently obtain noisy estimates of the quantity $\textbf{v}.$
They show that
if one knows the class imbalance, denoted by $b$ in their paper, then the specificities and sensitivities
have consistent estimators with errors of order $1/\sqrt{n}.$ 

Let us briefly consider the relation
between the finite sample estimate of the $\mu_i$ quantities to see their relation to the work
here.
\begin{align}
    \hat{\mu_i} = & \frac{1}{n} \sum_{s=1}^{n} f_i(x_s) \\
                = & \frac{1}{n} \suma (2 \indgt{i}{\lis{i}{s}} -1) + \frac{1}{n} \sumb (1 -2 \indgt{i}{\lis{i}{s}}) \\
                = & -1 + 2 \prva \left( \prsa{i} + \prsb{i} - 1 \right)
\end{align}
Note that this reformulation makes clear why the condition that the label accuracies must sum to a value greater
than one is not just to decode the true point. As in the work here, the spectral method approach also has blind spots.
The spectral blind spot is a line,
\begin{equation}
    \prsa{i} + \prsb{i} = 1.
\end{equation}

Finally, their solution for independent classifiers then finishes with their Theorem 1, a proof that a
restricted likelihood estimator for the class imbalance will converge to its true value in
probability as $n \rightarrow \infty.$ It is tempting to conclude that the independent evaluator is
much better because it provides sharp estimates that are exact for a finite sample. But this is not
a fair comparison. As was remarked in the section discussing the unresolved square root in the estimate
of \prva \, by the independent evaluator, the experiments carried here never observed a sample independent
test. A fairer comparison between the two methods would be one that considered the error in the independent
estimator for nearly sample independent ensembles versus the one provided by the spectral method for test
sizes where classifiers independent in distribution would produce similar sample correlations. That comparison
is not done here.

\section{The Platanios agreement equations and independent classifiers}

This section will discuss the mathematics of the Platanios agreement equations
(\cite{Platanios2014}) by mapping the algebra of agreement equations into the
algebraic evaluation framework presented here. We will see that while the general
agreement equations are correct, their application to the case of independent
classifiers is flawed and logically inconsistent. The detection of this theoretical
flaw in all the agreement equations work for independent classifiers is a
vivid demonstration of the main AI safety message of this paper - the nature of
algebraic solutions alerts to the failure of the evaluation model. In the case
of the Platanios agreement equations, the appearance of an unresolved square root
in their solution to the case of independent classifiers is that failure - no
purported solution to the case of independent classifiers can produce an estimate
for the integer ratio sample statistics that contains such unresolved square roots.

The approach of the agreement equations is to focus on a subset of the data sketch
presented here - just keep a record of how often members of an ensemble agree. In
the case of two binary classifiers that means that the agreement rate, $a_{i,j}$, between
the classifiers is just,
\begin{equation}
    a_{i,j} = f_{\alpha, \alpha} + f_{\beta, \beta}.
\end{equation}
In general, we can observe the agreement rate for any number of binary classifiers as,
\begin{equation}
    a_{i,j,k,\ldots} = f_{\alpha, \alpha, \alpha, \ldots} + f_{\beta, \beta, \beta, \ldots}.
\end{equation}
The $2^n$ data sketch values for $n$ binary classifiers are reduced to a single number. One
would immediately suspect that such a reduction would entail significant information loss.
And it does. The Platanios agreement equations are not able to solve for the full basic
set of sample statistics as done her. Instead, they solve for the stream average error 
rate ensembles of classifiers.  We can recast the stream average error rates in terms
of the basic evaluation sample statistics.

For a single binary classifier, the stream error rate can be written
in terms of the two label error rates of the classifier \emph{convolved} with the label
prevalences,
\begin{align}
    e_i& = \prva e_{i,\alpha} + \prvb e_{j,\beta} \\
     & = \prva (1 - \prsa{i}) + \prvb (1 - \prsb{i}) \\
     & = 1 - c_i,
\end{align}
where $c_i$ is the stream correctness rate for binary classifier $i$. The
expression for the stream correctness rate can be rewritten in terms of 
the basic evaluation statistics as,
\begin{equation}
    c_i = \prva \prsa{i} + \prvb \prsb{i}.
\end{equation}

We can write an expression for the agreement rate of pairs of
classifiers using the generating polynomials in the 
completeness theorem for the data sketch of correlated classifiers. For two
such classifiers we have four generating polynomials for the data sketch of
the pair's decisions,
\begin{align}
    \freqtwo{\alpha}{\alpha}& = \prva(\prsa{i} \prsa{j} + \corrtwo{i}{j}{\alpha}) + \prvb((1-\prsb{i})(1- \prsb{j}) + \corrtwo{i}{j}{\beta}) \\
    \freqtwo{\alpha}{\beta}& = \prva(\prsa{i} (1 - \prsa{j}) - \corrtwo{i}{j}{\alpha}) + \prvb((1-\prsb{i})\prsb{j} - \corrtwo{i}{j}{\beta}) \\
    \freqtwo{\beta}{\alpha}& = \prva((1 - \prsa{i}) \prsa{j} - \corrtwo{i}{j}{\alpha}) + \prvb(\prsb{i}(1- \prsb{j}) - \corrtwo{i}{j}{\beta}) \\
    \freqtwo{\beta}{\beta}& = \prva((1 - \prsa{i}) (1 - \prsa{j}) + \corrtwo{i}{j}{\alpha}) + \prvb(\prsb{i}\prsb{j} + \corrtwo{i}{j}{\beta}).
\end{align}
Combining the two polynomials for \freqtwo{\alpha}{\alpha} and \freqtwo{\beta}{\beta}, we have an expression for the agreement
rate of arbitrarily correlated pairs as,
\begin{equation}
    \label{eq:agrij-ae}
    a_{i,j} = \prva (\prsa{i} \prsa{j} + (1 - \prsa{i}) (1 - \prsa{j}) + 2 \corrtwo{i}{j}{\alpha}) + \prvb(\ldots).
\end{equation}
The expression times the \prvb \; term has the same algebraic structure as the \prva\; term, 
just the labels are switched.

We proceed by rewriting Eq \ref{eq:agrij-ae} in terms of either the correctness rate $c_i$ or the
error rate $e_i$,
\begin{align}
    a_{i,j}& = \prva ( 1 - \prsa{i} - \prsa{j} - \prsa{i} \prsa{j} + 2 \corrtwo{i}{j}{\alpha}) + \prvb(\ldots) \\
           & = \prva ( 1 - e_{i,\alpha} - e_{i,\alpha} - e_{i,\alpha} e_{j,\alpha} + 2 \corrtwo{i}{j}{\alpha}) + \prvb(\ldots).
\end{align}
The problem with the Platanios independent classifier solution now starts to become evident. By definition,
correlation term is an independent variable from that of the error rates of each classifier. So there is
no hidden algebra that obscures dependencies on the label accuracies or error rates or the classifiers.
The agreement rate is a convolution of these by-label terms and therefore cannot possibly decompose as claimed
erroneously in \cite{Platanios2014}. Let us carry out the math to see the problem.

\begin{align}
    a_{i,j}& = 1 - c_i - c_j + 2 ( \prva \prsa{i} \prsa{j} + \prvb \prsb{i} \prsb{j} + \prva \corrtwo{i}{j}{\alpha} + \prvb \corrtwo{i}{j}{\beta} ) \\
           & = 1 - e_i - e_j + 2 ( \prva e_{i,\alpha} e_{j,\alpha} \:  + \prvb e_{i,\beta} e_{j,\beta} \; \, + \prva \corrtwo{i}{j}{\alpha} + \prvb \corrtwo{i}{j}{\beta}).
\end{align}
Using the completeness theorem for correlated classifiers is a powerful analytic tool. It allows you to connect
events as logical statements - the joint stream error rate - to the algebra of the statistics.
\begin{align}
    c_{i,j}& = ( \prva \prsa{i} \prsa{j} + \prvb \prsb{i} \prsb{j} + \prva \corrtwo{i}{j}{\alpha} + \prvb \corrtwo{i}{j}{\beta} ) \\
    e_{i,j}& = ( \prva e_{i,\alpha} e_{j,\alpha} \:  + \prvb e_{i,\beta} e_{j,\beta} \; \, + \prva \corrtwo{i}{j}{\alpha} + \prvb \corrtwo{i}{j}{\beta}).
\end{align}
In other words, the algebra of the agreements and disagreements between noisy classifiers is not the same as the algebra
of the evaluation statistics. If they were, the problem of evaluation via algebraic methods on unlabeled data would be an
entirely solvable problem. This distinction between the two algebras is significant as it affects the theoretical claims
made in \cite{Platanios2014}. The paper confuses stream error independence with label independence. This is expressed
as a logical decomposition that is algebraically fallacious,
\begin{align}
    e_{i,j}& = e_i e_j \\
    ( \prva e_{i,\alpha} e_{j,\alpha} + \prvb e_{i,\beta} e_{j,\beta} )& = (\prva e_{i,\alpha} + \prvb e_{i,\beta}) (\prva e_{j,\alpha} + \prvb e_{j,\beta}).
\end{align}
The consequence of this that the solution for the error rate of independent classifiers is incorrect.

\subsection{Irrational evaluation estimates imply incorrect evaluation assumptions}

There is another way to see that the independent solution proposed by \cite{Platanios2014} cannot be
correct. The general idea of this paper is that finite evaluations have integer ratio ground truth
values. By using algebra properly, we can use this task postulate to detect failures of our evaluation
assumptions. In the main section of the paper, the counterpart to the algebraic evaluator was majority
voting. Its main AI safety drawback is that it always returns seemingly correct evaluations. In contrast,
the independent solution proposed by \cite{Platanios2014} will almost always return incorrect evaluation
estimates for independent classifiers. It is almost guaranteed to never return a correct evaluation.

It is instructive to go through the algebra of how one would prove that because it shows how this
theoretical analysis is so much easier than similar statements when probability theory is involved.
The solution proposed by \cite{Platanios2014} is,
\begin{equation}
    e_i = \frac{c \pm (1 - 2 a_{j,k})}{\pm 2 (1 - 2 a_{j,k})}.
\end{equation}
The difficulty arises in the expression `c' - it contains an unresolved square
root,
\begin{equation}
   c = \sqrt{(1 - 2 a_{1,2})(1 - 2 a_{1,3})(1 - 2 a_{2,3})}.
\end{equation}
By construction, for a finite test, the agreement rates between any set of classifiers
would be an integer ratio. So this term must resolve when the independence condition
applies because that is how this `c' term was derived. And it must resolve for \emph{any}
set of independent classifiers. That is an extraordinary coincidence any of us would be
hard pressed to accept. And we do not have to because the algebra of the derivation is
incorrect as we saw above. But that square root should have alerted the humans considering
this theory that something was wrong with it.

\section{General comments on the experiments}

All the code and data for the experiments is provided in accompanying files. The experiments were carried
out on a MacBook Pro with an M1 chip. To greatly simplify the comparison and to mimic artificial constraints
that may exist in production or field deployment the arbitrary decision was made that all the classifiers
would use a logistic regression method with pinned settings as provided in the Mathematica implementations
of the algorithm under their general function \texttt{Classify}. Similarly, the feature partitions used
for all the experiments where exactly the same - each classifier used just three features completely
disjoint from those used by the other ensemble members. In other words, no attempt was made to
show that the experimental protocol is optimal or the best possible choice for any of the datasets.

The timing for all the runs was dominated by training of the classifiers. Algebraic evaluation is
essentially immediate - one of its great benefits. The time taken to train 300 different feature partitions
10 times each over 10 different test sizes was about six hours.

\subsection{The distance to the containing variety}

The hypothesis that the independent evaluator estimate has non-zero distance to the containing
variety is only explored empirically in this paper. Given the small value of the distances
that are computed, one may wonder if this due to floating point errors. Mathematica has the
ability to compute this distance exactly when the surface is defined using integer ratios and
the independent estimate is given as an algebraic number. These exact calculations take considerably
longer than those done when floating point numbers are used. A few of these calculations were
carried out to confirm the correctness of the floating point estimates.

\subsection{Some evaluations}

It is instructive to consider the actual evaluations that were carried out using the feature
partitions in the experiments. The best evaluations were associated with the \texttt{twonorm}
experiments. Something that the plots of algebraic failures would lead us to expect. If
features are correlated, training protocols can only go so far in creating independent ensembles.
The \texttt{twonorm} dataset is synthetic and its features were independently generated. The
result of a single evaluation is shown in Figure~1. The evaluation was selected by performing
10 training/evaluation runs on the 100 feature partitions investigated in the 2nd set of experiments.
Those feature partitions that succeeded in giving seemingly correct values (all of them in the
\texttt{twonorm}) for the 10 runs were then searched for the one closest to the containing
variety for its evaluation.
\begin{figure}
  \centering
  \includegraphics[width=\textwidth]{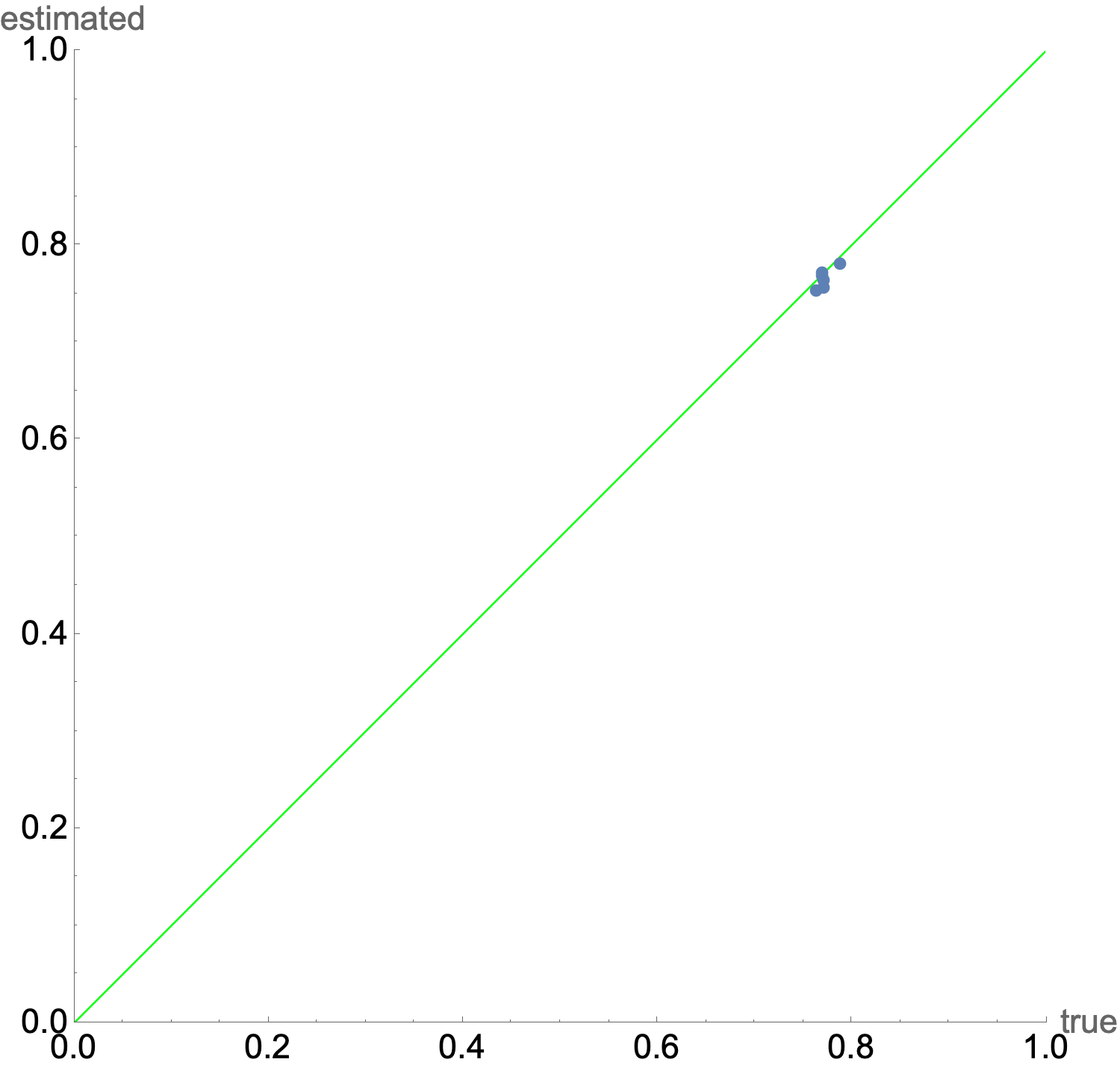}
  \caption{Estimated versus true values for the six label accuracies
  of three classifiers in a single evaluation of the \texttt{twonorm}
  features partition that had the closest distance to the containing variety.
  }
\end{figure}
The next best results for the minimum distance evaluation were obtained in the
\texttt{mushroom} dataset and are shown in Figure~2.
\begin{figure}
  \centering
  \includegraphics[width=\textwidth]{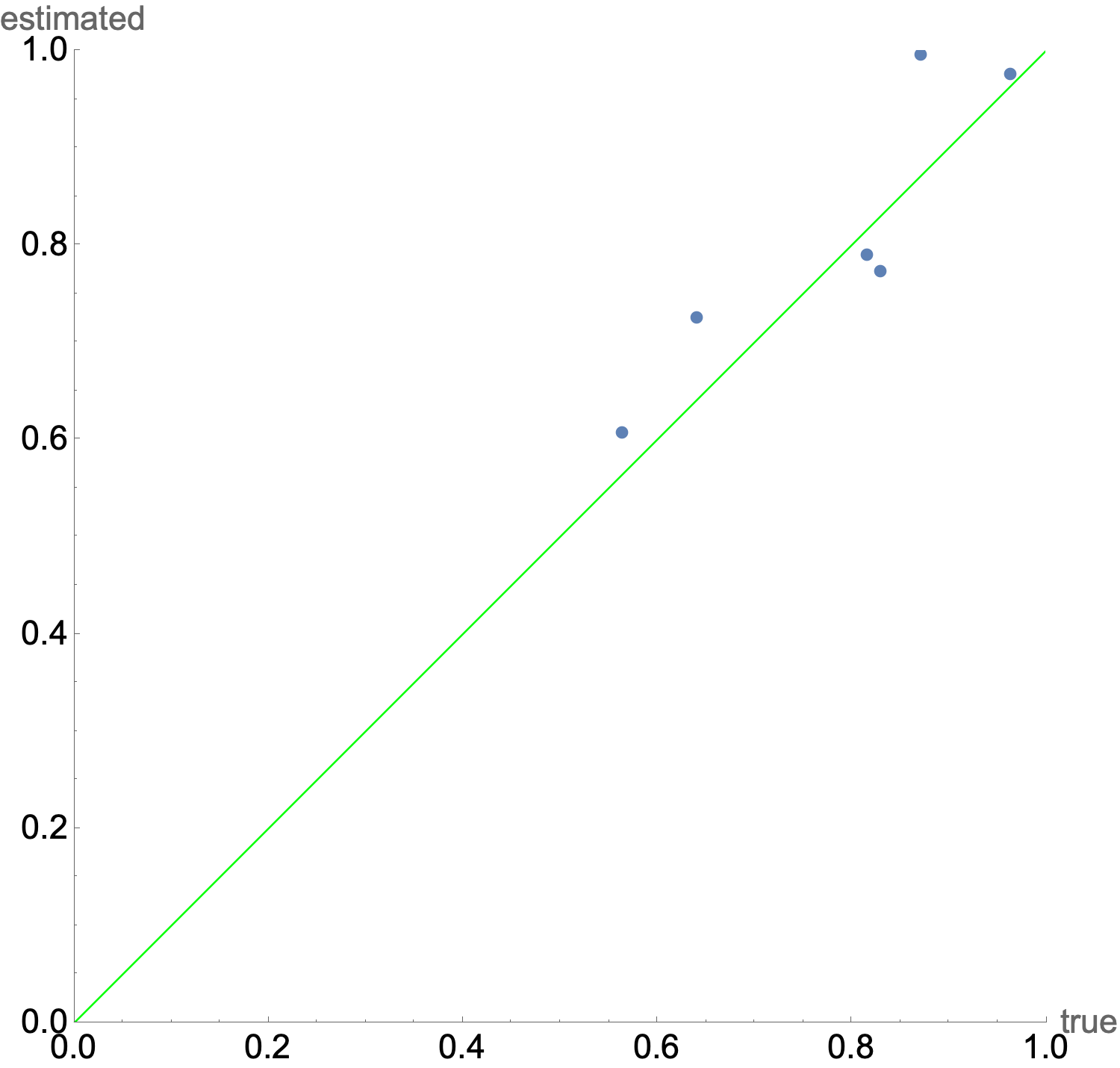}
  \caption{Estimated versus true values for the six label accuracies
  of three classifiers in a single evaluation of the \texttt{mushroom}
  features partition that had the closest distance to the containing variety.
  }
\end{figure}
And finally, the \texttt{adult} evaluation was the worst.
\begin{figure}
  \centering
  \includegraphics[width=\textwidth]{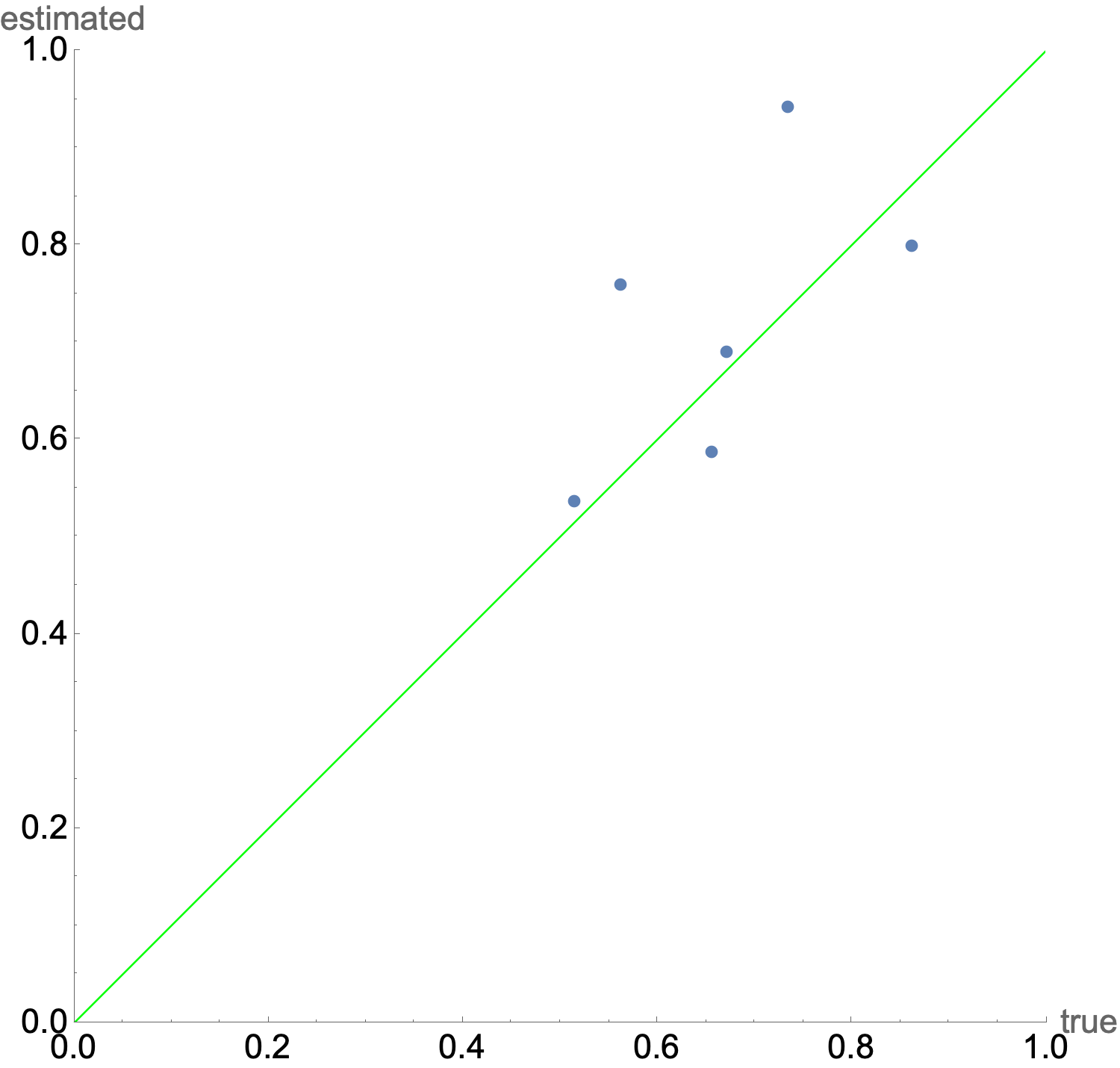}
  \caption{Estimated versus true values for the six label accuracies
  of three classifiers in a single evaluation of the \texttt{mushroom}
  features partition that had the closest distance to the containing variety.
  }
\end{figure}

If one is concerned with mitigating the principal/agent monitoring paradox,
exhaustive searches of evaluating ensembles are hardly practical and are,
therefore, not the focus of this paper. Much more practical when one is concerned
about safe or profitable deployments is being able to handle evaluations were
the classifiers are nearly independent. The algebraic approach presented here
is a step in that direction but work remains on handling these cases. Knowing
evaluations failed is hardly useful when that is the common case.

\section{References}

\bibliographystyle{unsrtnat}
\bibliography{streamalgeval}